\documentclass{article}
\usepackage{arxiv}
\usepackage{alphabeta}
\usepackage{subcaption}
\DeclareUnicodeCharacter{0308}{HERE!HERE!}
\DeclareUnicodeCharacter{0220}{HERE!HERE!}%\inputencoding{ascii}
\usepackage{amsmath}
\usepackage{hyperref}
\usepackage{amsfonts}
\usepackage{multirow}
\usepackage[T1]{fontenc}    % use 8-bit T1 fonts
\usepackage{xspace}
\usepackage{graphicx}
\usepackage{adjustbox}
\usepackage{float}
\usepackage{booktabs}
\usepackage{amsfonts}       % blackboard math symbols
\usepackage{nicefrac}       % compact symbols for 1/2, etc.
\usepackage{microtype}      % microtypography
\usepackage{lipsum} 
\usepackage[normalem]{ulem}
\usepackage{hyperref}
\usepackage{fancyhdr}       % header
\usepackage{graphicx}       % graphics
\graphicspath{{media/}}     % organize your images and other figures under media/ folder
\usepackage[affil-it]{authblk}
\usepackage{adjustbox}
\usepackage{amsmath}
\usepackage[dvipsnames,table,xcdraw]{xcolor}
\usepackage{hyperref}
\usepackage{xcolor}
\usepackage[font=scriptsize]{caption}
\usepackage{mwe}
\usepackage{longtable,pdflscape,booktabs}
\usepackage{textcomp}
\usepackage{multirow}
\usepackage{longtable,pdflscape,booktabs}
\usepackage{textcomp}
\usepackage{tabularx}
\makeatletter
\usepackage{makecell}
\usepackage{amsmath}
\usepackage{array}
\usepackage{fancyhdr}
\DeclareRobustCommand{\greektext}{%
  \fontencoding{LGR}\selectfont\def\encodingdefault{LGR}}
\DeclareRobustCommand{\textgreek}[1]{\leavevmode{\greektext #1}}
\ProvideTextCommand{\~}{LGR}[1]{\char126#1}

\providecommand{\tabularnewline}{\\}
\makeatother

\newcolumntype{P}[1]{>{\centering\arraybackslash}p{#1}}
\newcolumntype{M}[1]{>{\centering\arraybackslash}m{#1}}
 \title{Artificial Intelligence in Material Engineering}

\author[1]{Lipichanda Goswami}%\thanks{These authors contributed equally to this study}
\author[1]{Manoj kumar Deka}

\author[2,*]{Mohendra Roy} %\footnote[2] \printfnsymbol{3}}

\affil[1]{%
  Department of Computer Science and Technology, Bodoland University, BTC, Assam, India}
\affil[2]{%
  Department of Information and Communication Technology, Pandit Deendayal Energy University, Gandhinagar 382007, India
  }

\affil[*]{%
  \textnormal{Corresponding author: mohendra.roy@ieee.org }}
 
\begin{document}
\maketitle
%\pagestyle{fancy}
% \rhead{\includegraphics[width=2.5cm]{vch-logo.png}}

\keywords{Artificial Intelligence, Density Functional Theory, Material Engineering, Deep Learning, Graph Neural Network,
\newline \textcolor{red}{\textbf{Note:} This article is available in the Journal of Advanced Engineering Materials from 11 April 2023. } }
%\fancyfoot[R]{Copyright: Wiley: https://onlinelibrary.wiley.com/doi/10.1002/adem.202300104}
\fancyfoot[R]{Page \thepage \hspace{1pt} }

% Abstract should be written in the present tense and impersonal style (i.e., avoid we), and be at most 200 words long

\cfoot{\text Copyright: Adv. Eng. Mater. 2023, https://onlinelibrary.wiley.com/doi/10.1002/adem.202300104}

\begin{abstract}

\textbf{Abstract:} The role of artificial intelligence (AI) in material science and engineering (MSE) is becoming increasingly important as AI technology advances. The development of high-performance computing has made it possible to test deep learning (DL) models with significant parameters, providing an opportunity to overcome the limitation of traditional computational methods, such as density functional  theory (DFT), in property prediction. Machine Learning (ML) based methods are faster and more accurate than DFT based methods. Furthermore, the generative adversarial networks (GANs ) has facilitated the generation of chemical compositions of inorganic materials without using crystal structure information. These developments have significantly impacted the material engineering (ME) and research. Some of the latest developments in AI in ME herein are reviewed. First, the development of AI in the critical areas of ME, such as in material processing, the study of structure and material property, and measuring the performance of materials in various aspects, is discussed. Then, the significant methods of AI and their uses in MSE, such as graph neural network, generative models, transfer of learning etc are discussed. The use of AI to analyze the results from existing analytical instruments is also discussed. Finally, AI’s advantages, disadvantages, and future in ME are discussed.

\end{abstract}

% Text: Please use section headings and subheadings as specified below. For communications, all section headings apart from Experimental Section should be removed
% Please make the first reference to a display item bold: \textbf{Figure 1}
% Do not abbreviate Figure, Equation, etc.; display items are always singular, i.e., Figure 1 and 2.
% Equations are always singular, i.e., Equation 1 and 2, and should be inserted using the {equation} environment, not as graphics
% Please do not use footnotes in the text, additional information can be added to the Reference list.
\section{Introduction:} 

Material Science and Engineering (MSE) is mainly concerned with four characteristics of a material. These are processing, structure, property, and performance. The key to material engineering lies in the interrelation of these four characteristics. In short, the combinations of processing, structure, property, and performance are key in material engineering.\cite{1} Here the structure represents the atomic arrangements of the material. Performance defines how well the material plays its role in a particular task. Properties like hardness/softness, the density of the particles, fracture toughness, resistivity, and thermal expansion are determined by the structure. These properties can be engineered by adopting appropriate processing methods. Here, the processing is a series of steps that are involved to convert a material to some useful form by tweaking the properties of the material. Engineered materials such as metals, polymers, liquid crystals, and composites are widely used in the fields such as medicine, energy, manufacturing, biotechnology, etc. Therefore, MSE is an emerging area applicable in a variety of materials for multiple disciples like medical science, biotechnology, nanotechnology, drug discovery, energy storage materials, etc.

Some of the traditional techniques used in MSE are i) Density Functional Theory (DFT) which is a simulation method that uses the quantum mechanical laws to find out the electrical properties of atoms, molecules, and solids. \cite{6}. ii) Density Functional Perturbation Theory (DFPT) where the quantum system is studied in small perturbation mostly used in calculating vibrational energies of phonon that can further be used to find out the physical properties. \cite{7} iii) Classical force-field inspired descriptors (CFID)that represent the chemistry-structure-charge data of a material.
But these traditional methods take a considerable amount of time in processing and analysis. Also, these can not be applied in all types of structures\cite{75}.

With years of study and experimentation on the conventional methods of property prediction, such as the empirical trial-error method, and density functional theory (DFT), researchers have collected huge data in the field of MSE. These Big data may help in designing a data-driven approach for MSE. In this regard, Artificial Intelligence (AI) can play a major role. AI is an area of computer science that leads a system to learn from data and improves performance in every subsequent iteration.  The learning process starts with the observation of data to find out the meaningful features to attain the set objectives. In the last few years, with the increasing experimental and simulation-based dataset, AI and machine learning(ML) have been widely used to gain a deeper insight into the material.

\subsection{A Brief discussion on existing literature on AI in material engineering:}

A handful of reviews have already been made in the context of applications of AI in material engineering. For example, Kamal Choudhary et al. have discussed the available ML techniques and their libraries \cite{2}, Chi Chen et al. have discussed the ML methods that are specifically used for energy materials \cite{3}, Jonathan Schmid et al. have discussed the ML algorithms for crystal structure prediction \cite{4} and Valentin Stanev et al. have discussed the AI models that are used in quantum materials.\cite{5}. Daniel P Tabor and co-authors provided a fruitful discussion on discoveries in the clean energy sector along with the state-of-the-art procedures for organic, inorganic, and nanomaterials \cite{40}.
Table \ref{table4} of the supplementary document contains the details of models that are built particularly in the field of organic, inorganic, energy storage, drug and pharmaceutical, and biomaterials. In the same work, high throughput virtual screening, genetic algorithms in the synthesis of catalysts, and ML algorithms in perovskite synthesis for photovoltaic are discussed. The challenges that are still found in the automatic synthesis of inorganic and organic molecules, the autonomous laboratories fueled by AI models for chemical synthesis, the techniques used for automatic and rapid characterization of materials, the use of autonomous robots in the laboratories for speeding up the experimentation are also mentioned with state-of-the-art works. The use of generative and discriminative neural networks in photonic devices, the types of datasets particularly used for electromagnetics, and some of the dimensionality reduction techniques are elaborately discussed by Jiaqi Jiang et al.\cite{41}. Mohit Pandey et al. have discussed the use of Graphics Processing Units (GPU), Deep Generative Networks, and transfer learning models have shown significant acceleration in the field of drug discovery \cite{42}. The GPU-based systems can reduce the computational cost to a great extent in comparison to the CPU in the simulation of molecular dynamics. GPU-specific quantum chemistry codes such as TeraChem are developed for simulating the entire protein structure using DFT.
 
  \subsection {AI in structural, elemental, electronic and thermal, di-electric and mechanical property prediction:}

\begin{figure}[!h]
 \centering
\includegraphics[width=.6\linewidth]{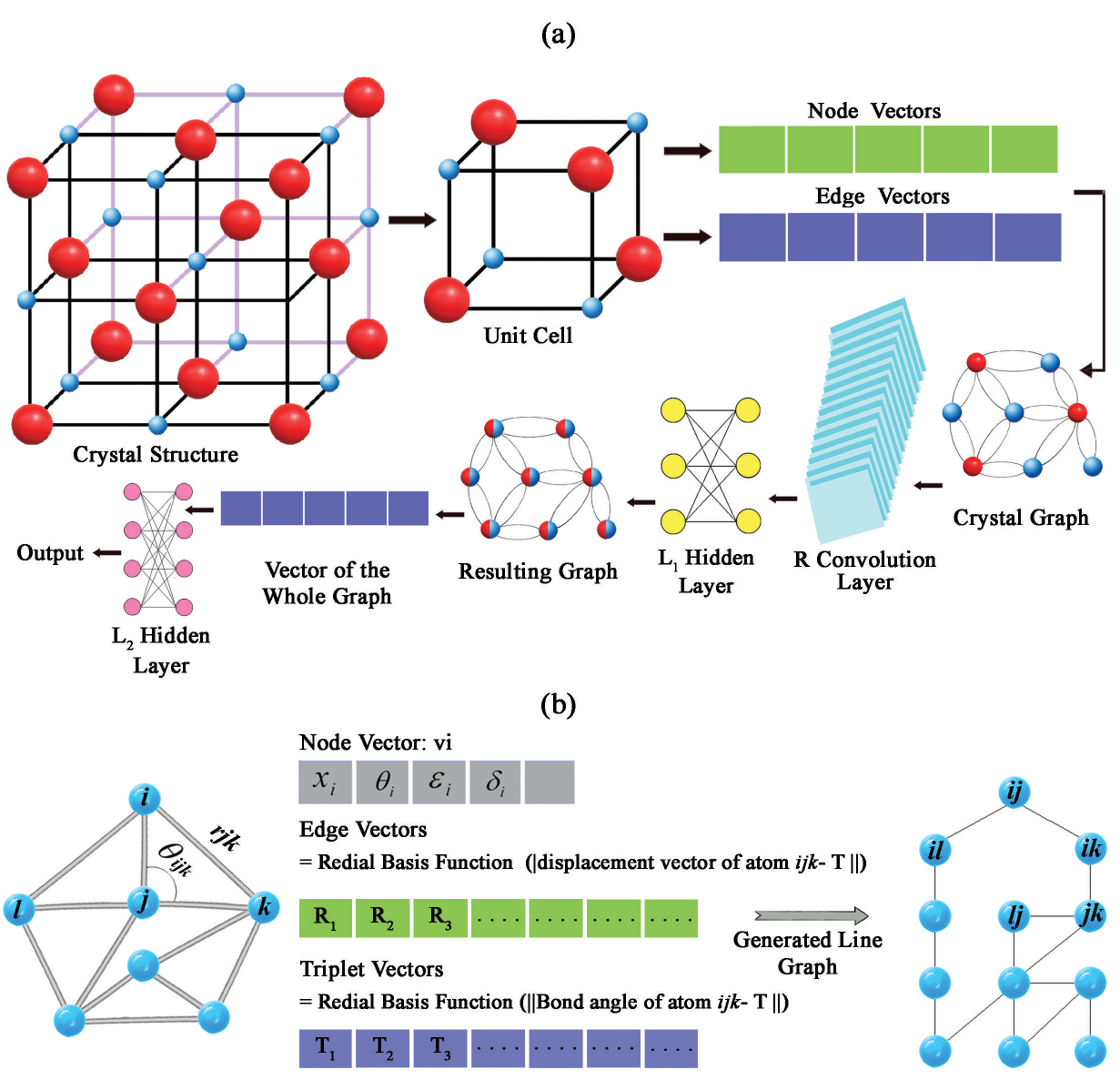}
\caption{Two of the mostly used Graph Neural Networks in the literature. \textbf{a)}Schematic of the CGCNN model: The crystal structure is converted to a crystal graph by taking atoms as nodes and atomic bonds as edges from the unit cell. The nodes of the graph go through R convolution layers and L1 hidden layers to produce a resulting graph that considers the local environment of each of the atoms. After the pooling layer, a vector of the whole graph is produced and sent to L2 hidden layer for further processing. The L2 layer then produces the predicted property as output. \newline \textbf{b)} Schematic of the ALIGNN model. The graph on the left is the bond graph of a crystal structure. The nodes of the graph are analogous to the atoms, and the edges are analogous to interatomic bonds. From this graph, another graph, L(g) (right), known as the line graph, is derived by considering the edges of the bond graph as nodes and the interatomic bond pair or the triplet of atoms as edges. Message passing is performed between the bond graph and the line graph in the convolution layer.}
\label{fig1}
\end{figure}

 Extending the existing review of AI in MSE, we focused on AI and ML methods that are developed for various material types like organic, inorganic, energy-storage, bio, and pharmaceutical materials to predict Stoichiometric, electronic, elemental, ionic, optical, structural, and thermal properties.
Recently, the Scanning Tunneling Microscope (STM) images have been used in convolutional neural networks (JARVIS-STMNet) to classify the structure of Bravais-lattices. This classification is used for phase identification, information extraction from poor resolution, etc. \cite{8}. ML models designed using Gradient Boosted Trees (GBT) algorithm has found to be superior as compared to many other classification models in predicting the topological structure of materials .\cite {21}. The famous Random Forest algorithm is used to find out the critical temperature (Tc) value of superconductivity.\cite{22}
Again graph neural network named Atomistic Line Graph Neural Network (ALIGNN) is used by Kamal Choudhary and his team to predict the structural and electronic properties and also for the quantities like adsorption isotherm of $ CO_2$ for various pressure\cite{15}. A schematic of the ALIGNN is shown in Fig \ref{fig1}. The same model is used to train the Density of Space Spectra (DOS) in two different representations, such as discretized representation and a low-dimensional representation of the crystalline materials, by training two models, AE-ALIGNN and D-ALIGNN separately.\cite{24} The derived DOS is helpful in gaining a deeper insight into the electronic properties of the materials and their relationship with ingredient species. 
For the electronic property prediction, a recent deep learning (DL) model ElemNet is used by leveraging the concept of transfer learning (see the transfer learning architecture in figure \ref{fig3}  ). The stability of a compound (by predicting the formation energy) could be successfully determined for both DFT computed dataset and the experimental-based smaller dataset. The model is also applicable in predicting thermal, mechanical, and magnetic properties, which are expensive to calculate experimentally.\cite{9} In a modified version of the ElemNet model, the concept of cross-property deep transfer learning is incorporated for predicting the electronic properties.\cite{11}. The Localized Gaussian Process Regression (L-GPR) model that uses a smaller dataset is applied to screen the materials based on formation energy.\cite{19}  For the same task, Crystal Graph Convolutional Network (CGNN) with slight modification in the convolutional layer has shown improved accuracy compared to DFT-based methods. The modification is done by ignoring the difference in interaction strength between neighbor nodes in the convolutional layer of the network. The same framework can also predict electrical and physical properties with improved accuracy.\cite{28} Again, Embedding the concept of multi-task learning in the CGCNN model, the relative stability of different materials based on their formation energy and classification task like metal/non-metal based on the acquired bandgap is carried out by Soumya Sanya et al. \cite{46}.
Another recent model developed by Mohammadreza Karamad et al. \cite{39} is the Orbital Graph Convolutional Neural Network (OGCNN) that uses the orbital field matrix (OFM) descriptor, a data representation method taken from the one-hot vector concept of natural language processing. Here, the representation of an atom is embedded with the orbital-orbital interaction of atoms and the long-range interactions in the local structure. The incorporation of OFM descriptor in the GCNN has improved prediction accuracy for electronic properties like bandgap, Fermi Energy, and Formation energy as compared to state-of-the-art GCNN models.
A description of these models is given in detail in Table \ref{table2} of Supplementary materials.
In \cite{51}, the existing SchNet model is extended by including an edge update network because of which the hidden state of receiving atom is responsible for the information interchange among the atoms. This extension improved the accuracy in formation energy prediction. Atom2Vec model is designed for the prediction of the formation energy of elpasolite crystals that are used for radiation detection.\cite{34} The CGCNN and Materials Graph Neural Network (MegNet) are used in predicting magnetic moment and formation energy. Though the models performed well in predicting formation energy, they showed poor performance in the magnetization of data.\cite{53}
For bandgap prediction of a class of complex oxide, Double perovskite structure, a statistical learning model, Kernel ridge regression (KRR), is cross-validated by the Linear Least Square fit (LLSF) model. These models have given more profound insights into double perovskite structure.\cite{20} But the KRR model is limited to a considerable chemical space and the nonmagnetic perovskite $AA'BB'O_6$.
DARWIN: Deep Adaptive Regressive Weighted Intelligent Network designed in the work of \cite{50} uses Graph Convolutional Neural Network(GCNN) with some edge attributes for learning from a smaller dataset. The model has been successfully used to predict UV materials' bandgap and the energy to estimate structural stability.
 The DeeperGATGNN is designed using a global attention graph neural network with the inclusion of residual skip connection and differentiable group normalization to allow the network to go deeper. The model showed outstanding performance in bandgap prediction by increasing the hidden layers above 20, whereas other state-of-the-art models tend to crash with more hidden layers. It is also free from overfitting.\cite{60}
 In addition to property prediction, uncertainty evaluation of ML models in order to determine the trustworthiness of computed data is carried out in \cite{14}. Here three different approaches are used. The first one is the GBDT method with quantile loss function, defined as in equation (1). \begin{equation}
     L(x_i^p, x_i)=max[q(x_i,x_i^p),(q-1)*(x_i-x_i^p)]
      \end{equation}
      where q is the quantile that gives a value to a group for the observations to fall within that value. \begin{math}  x_i^p \end{math} is the prediction and \begin{math} x_i \end{math} is the outcome. In the second approach, GBDT is used as a base model for the prediction of property, and Gaussian Processes (GP) is used as an error model for finding the prediction intervals. In the third approach, GP is used for determining the uncertainty of the trained model. These models are used for electrical, energetic, mechanical, and optical properties, and the obtained results are compared with each other.
      The recent progress of ML models has paved the way for solving a basic problem of quantum mechanics: the Schrodinger equation (SE), defined as 
\begin{equation}
    H \psi = E \psi
\end{equation}
The Kernel Ridge Regression Model is used in this task that can establish a non-linear relationship between the atomization energy of a molecule and its characteristic, thereby solving the molecular SE.
The trained model is successful in finding new molecular systems with different geometry and composition.\cite{23}
The concept of the descriptor 'bag of words, used in natural language processing, performs the encoding of the frequency of a particular word in a text, which is mimicked in the work of Hansen et al. \cite{25} by encoding the interatomic distances in Bag of Bond (BoB) descriptor in the chemical compound space. The model performed well in atomization energy and total energy prediction of molecules.
The ALIGNN model in \cite{12} is modified by incorporating the angular information of the atoms in the line graph to get a better atomic structure and is applied for various electronic and molecular properties of the DFT computed dataset.\cite{12}. 

\break Molecular properties are also predicted in \cite{33} by  Hierarchically Interacting Particle Neural Network (HIP-NN) that uses Linear regression for modeling the local hierarchical energies and Adaptive Moment Estimation (Adam) algorithm for training the model.

 For the prediction of dielectric and mechanical properties like Born-effective charge (BEC) tensor, piezoelectric (PZ) tensor, and IR frequency, a research group in \cite{18} have used structural descriptors such as Classical force-field inspired descriptors (CFID) and ML models based on gradient-boosting decision tree (GBDT) for both classification and regression task. The regression model yields good accuracy in finding the highest infrared frequency and maximum BEC, whereas the classification models are used for classifying high PZ and dielectric materials. (CFIDs) and (GBDT) are also used to speed up the screening process in predicting magnetic properties of materials.\cite{13}
 A recently developed model, deeper graph neural networks (deeperGATGNN), that uses ResNet structure and differentiable group normalization in the graph attention layers can learn the relationship between crystal structure and their vibrational energy. The accuracy of the model was suitable for rhombohedral crystals, but it reported high MAE for cubic structures while training on mixed samples revealing the low structural transferability of the model.\cite{16}

  %======= Energy Material============

 \subsection{AI in energy material engineering:}
  
 A handful of works in energy materials have also been carried out in recent years. Linear regression, Reduced error pruning (REP) tree, Rotation forest+REP tree, and Random subspace+REP tree are explored for predicting band gap of solar cells and metallic glass alloys, giving remarkable accuracy. In this work, a diverse set of descriptors are generated by creating Stoichiometric attributes, electronic structure attributes, elemental property statistics, and Ionic compound attributes.\cite{26} The Gradient Boosting Decision Tree (GBDT) is also used as a binary classification model to find out the promising solar absorber materials by classifying the data based on spectroscopic limited maximum efficiency (SLME) for quick pre-screening of the materials.\cite{10}
  For screening the high-pressure alloys that are used for hydrogen storage,  ML models like RepTree, RFR, and Neural Networks are used. These models are trained to predict the thermodynamic properties, such as hydride enthalpies and entropy of hydrogenation of the alloys.\cite{17} In the same work, low-energy Ti–Mn–Fe structures are detected by predicting the structure and phase with the help of a genetic algorithm. However, while comparing this structure with DFT and CALPHAD studies, contradictory results are found. In \cite{49}, a dataset for LiSi battery materials is designed by finding the random structure relaxations of the material using DFT. 
  The datasets used in the state-of-the-art models are described in table \ref{table3} of supplementary materials.
  For measuring the quantity of dendrite growth in the initiation phase of Li metal anode, ML frameworks have been used recently.\cite{57} In this task, different ML models are used in different phases. In the Isotropic material screening phase, where the stability of electrodeposition is determined for solid electrolytes, the shear and bulk moduli are calculated using the crystal graph convolutional neural network (CGCNN), and in the second phase, that is, anisotropic material screening, AdaBoost, Lasso, and Bayesian ridge regression are used for prediction of
\begin{math}C_{11} \end{math}, \begin{math}C_{12} \end{math} and \begin{math}C_{44} \end{math} 
elastic constants. \newline The Open Catalyst 2020 (OC20), a DFT-based catalyst dataset, is designed by L Chanussot et al. \cite{64} that contains surfaces and adsorbates for renewable energy storage. The dataset is used to train deep learning models such as CGCNN, SchNet, and DimeNet++ for random structure relaxation, Structure to Energy and Forces (S2EF) prediction, and relax state energy prediction from an initial structure. In the field of renewable energy materials, researchers have worked to develop suitable ML models for alkaline ion batteries like electrodes, electrolytes, photovoltaic materials, catalytic materials, etc.\cite{67} A unified statistical model for potential energy estimation named PreFerred Potential (PFP) is designed to control any combination of 45 elements of the periodic table and applied for a variety of tasks such as calculating the activation energy of battery materials without including the material in the training dataset to optimize the crystal structure of MOF and in finding the transition temperature of Cu-Au alloys.\cite{66}
In thermoelectric property prediction, Laugier et al. \cite{Laugier} tried the CGCNN model but was not successful in bypassing DFT. In this work, a Fully Connected Neural Network (FCNN) is trained on thermoelectric power factor obtained from DFT + Boltzmann Transport Equations (BTE) and found that though the convergence time is more in FCNN, it showed good prediction accuracy.
A modified version of CGCNN, where the local environments of each crystal are represented by vectors that are dependent on the composition and structure and independent of human-designed features, is used in perovskites, elemental Boron, and inorganic crystals. This model is capable of predicting Electronegativity, group number, radius, and element blocks of perovskites with remarkable accuracy. Also, complex Boron configurations are successfully explored by the model. But in local energy prediction of inorganic materials, it showed poor performance\cite{54}
The recent progress in ML models has shown great potential in predicting multi-property in a single model. Materials Graph Network (MEGNet), an upgraded version of Graph Neural Network, is used for predicting internal energy for multiple temperatures, enthalpy, entropy, and Gibbs free energy with temperature within the same model.\cite{29} The model also showed good accuracy in electrical property prediction. An important strategy included here is the unification of multiple free energy MEGNet models into a single one by integrating global state variables such as pressure, entropy, and temperature.

 \begin{figure}[!h]
 \centering
\includegraphics[width=.7\linewidth]{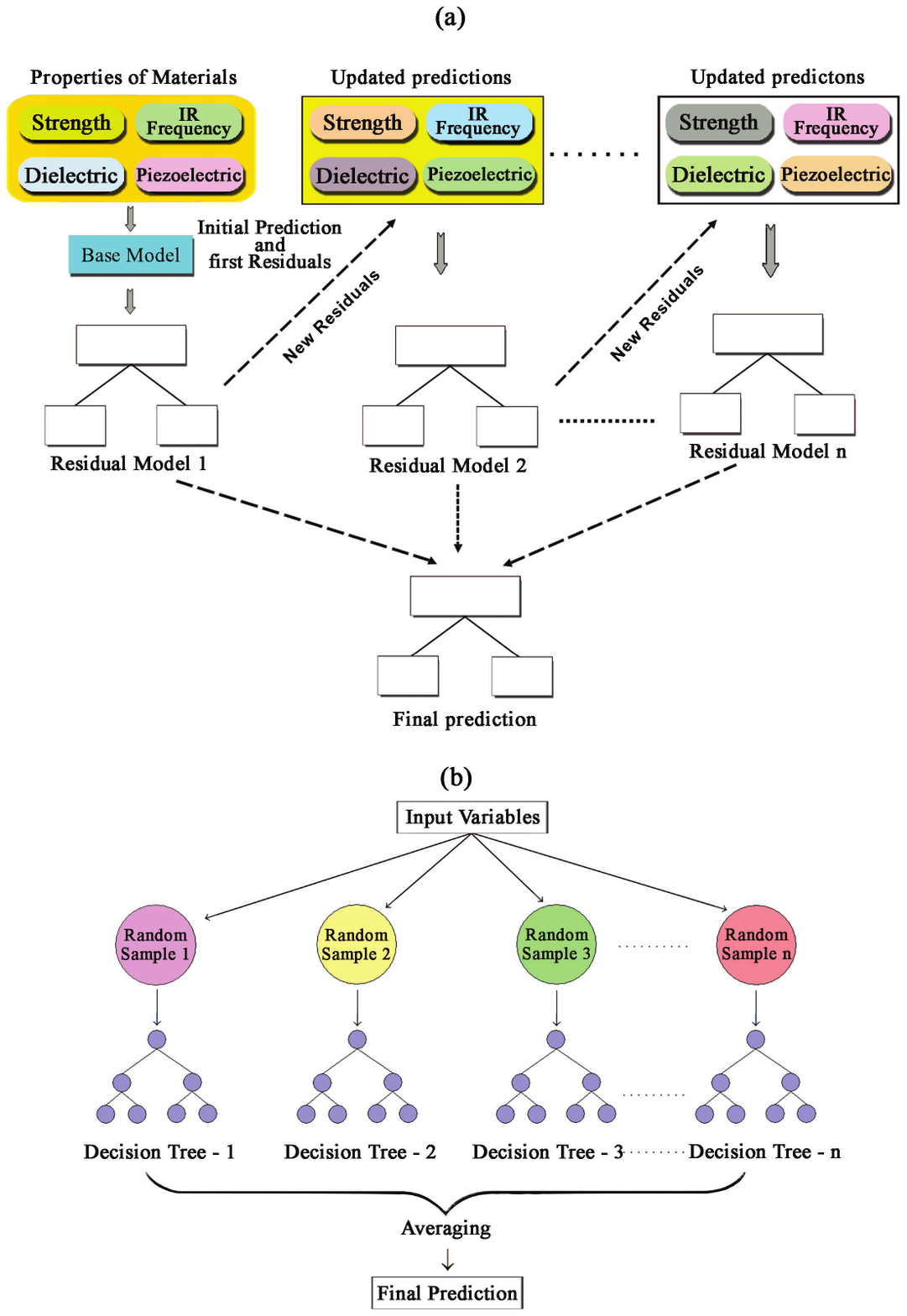}
\caption{Two frequently used Machine Learning MOdels \textbf{a)}Schematic of the Gradient Boosted Decision tree regression model: From a given set of data, a base model finds out the initial prediction and sets the first residuals. The residual is sent to a weak model known as a residual model. Each of the residual models updates its prediction by the new residuals, and finally, a strong model, which is an ensemble of all the weak models, gives the final prediction value.
\newline \textbf{b)}Schematic of a Random Forest Regression (RFR). From a set of Input Variables, random samples are fed to a set of decision trees. The output generated by the trees is averaged and considered as a final result.}
\label{fig2}
\end{figure}

%\break 

Because of this advancement, the model is capable of predicting many important properties like internal energy for multiple temperatures, enthalpy, entropy, and Gibbs free energy with temperature and can eliminate the need to design a separate model for each of the properties. Another multi-task learning model HydraGNN developed by M L Pasini et al. \cite{30}, is capable of predicting atomic magnetic moment, mixing enthalpy and atomic charge transfer at the same time. The model architecture consists of two sets of layers, the first set of layers learns the standard features, and the second set learns the specific features of any material property. In the convolutional layers of HydraGNN, a variation of the Graph convolutional neural network (GCNN) known as Principal Neighbourhood Aggregation (PNA) is included, which makes it easier for the model to classify two different graphs.
ML is fueling the growth of polymer industries and nanoparticles also. In polymer laboratories, AI-powered autonomous robots are developed that are capable of performing structure-function testing.\cite{44} 
In nanoparticle synthesis, stable noisy optimization by branch and fit (SNOBFIT) algorithm, Microfluidic, and Robotic based systems integrated with ML techniques have gained importance. Whereas Neural Networks are used in carbon nano particles.\cite{45}
In \cite{52}, a CNN with a U-net architecture is used for finding the region of interest by performing segmentation on high-resolution transmission electron microscopy (HRTEM). Here, defect detection for individual nanoparticle regions is performed by random forest (see figure \ref{fig2} for Random forest architecture). Other significant works in numerous fields include: feedforward neural network that is used in X-Ray diffraction (XRD) patterns to solve the space group determination problem\cite{31},Teacher-student deep neural networks (TSDNN) for new stable materials with negative formation energy and synthesizability for both labelled and unlabelled data with better accuracy than prior models\cite{61}, Optimization of semiconductor dot devices using ML algorithms\cite{43}, supervised ML models in interlayer energy and elastic constant prediction of essential heterostructure for solid and super lubricant materials\cite{62}, Deep Potential Generator (DP-GEN) framework for statistical, mechanical and dynamical properties prediction of Al, Mg and Al-Mg alloys and for the modelling of Potential Energy Surface (PES) using the molecular dynamics (MD) simulation and without inclusion of structural information\cite{48}, 3-D chemical structure prediction using Variational Autoencoder and a 3-D U-Net segmentation network with an attention mechanism.\cite{56}, prediction of compounds having highest melting temperature using ordinary least squares regression (OLSR), partial least-squares regression (PLSR), support vector regression (SVR),and Gaussian process regression (GPR) models by preparing separate predictor sets, where first set contains the physical properties and the second one contains both physical and elemental properties\cite{27}
%======== AI in molecular Engineering 

\subsection{AI in predicting molecular property:}

In molecular property prediction, frameworks like MoleculeNet are designed for predicting four categories of molecular properties: quantum mechanics, physical chemistry, biophysics, and physiology by preparing separate datasets for each of these properties and creating a separate metric and splitting pattern for each of these datasets.\cite{36} Graph Convolutional Networks (GCN) and Graph Isomorphism Networks (GIN) with self-supervised learning mechanisms in order to improve the classification and regression task are used to design the framework Molecular Contrastive Learning of Representations (MoLCLR) for molecular property prediction.\cite{37} The deep tensor neural network (DTNN) designed for quantum chemical property prediction is scalable according to the number of atoms in a molecule. It is capable of predicting data beyond the training set. The DTNN is successfully used in isomer energy prediction, in the classification of molecules based on carbon ring stability, and in some peculiar electronic structure prediction giving uniform accuracy for intermediate-size molecules.\cite{58}  The KV-PLM model is capable of learning the co-relation between biomedical text and molecular structures and hence can assist the discovery of the drug.\cite{35} The Schnet model designed to search the chemical space and energy surfaces was successfully used for predicting quantum mechanical property for \begin{math} C_{20} \end{math} -fullerene, which was not possible with the simulation method.\cite{59} For predicting physicochemical properties from molecular structures, a multiplex graph neural network named Multiplex Molecular Graph Neural Network (MMGNN) is designed and found to be efficient.\cite{32}
Beyond the concept of property prediction for some existing materials, an advanced concept in material science is the inverse design framework. Wherein new materials are discovered for a given target property. This is manifested by a deep variational autoencoder neural network with a supervised mechanism called active learning and a generative adversarial deep neural network and have discovered eleven different semiconductor materials and two materials with high bandgap. \cite{47}. In \cite{55}, a high-speed inverse design framework named as Fourier-Transformed Crystal Properties (FTCP) for inorganic crystals that can predict the structure and chemistry of a material upon giving some targeted property is designed. The framework is then used for targeted formation energies, bandgap, and for thermoelectric power factor. Conditional GAN (CondGAN) and conditional VAE (CondVAE) models are used to implement the inverse design concept where inorganic compositions are generated from the target property without including the crystal information.\cite{63} Various machine Learning models are also used for engineering new concrete formulas for desirable properties.\cite{83} The realistic samples are generated by conditional distribution $p(y|x)$ so that the properties of the generated data can be controlled by changing the value of $x$. Where $x$ contains the information about strength, age, and environmental impact, and $y$ represents the number of constituent materials. The reduced environmental effect and strength of the formulas are then verified using a regression model by estimating the similarity between generated properties with the desired properties.
 Some of the tools and frameworks used in implementing DL and ML algorithms in state-of-the-art works are described in table \ref{table1}

\begin{longtable}{p{2.5cm}|p{6cm}|p{6.5cm} } 
%\rowcolors{1}{gray!40!white}{blue!40!white!80} 
\caption{Tools and Frameworks}
\label{table1}\\[0.5ex]
\hline
   \textbf{Tools and Frameworks} &	\textbf{Description} &	\textbf{Link}
 \\[0.5ex]
 \hline
 Scikit-learn	& An Open source easy to use and efficient tool for predictive data analysis 	& https://scikit-learn.org/stable/ \\[0.5ex]
 \hline
 Tensorflow	& A free and open source library specifically designed for DL models & https://www.tensorflow.org/ \\[0.5ex]
 \hline
Theano	& The Python library for DL that can perform fast numerical computations that includes multi-dimensional arrays	& https://pypi.org/project/Theano/\\[0.5ex]
 \hline
Caffe	& A deep learning framework having an expression architecture that allows the users to switch between CPU and GPU	& https://caffe.berkeleyvision.org/installation.html
\\[0.5ex]
 \hline
MXNet	 & It is a fast and scalable DL framework that allows training of fast model and supports multiple programming language.	& https://mxnet.apache.org/versions/1.9.0/ \\[0.5ex]
 \hline
Keras	& An open source library that acts as an interface for the tensorflow library	& https://keras.io/ \\[0.5ex]
 \hline
Pytorch	& Pytorch is a python package that gives two features tensor computation with strong acceleration by GPU and deep neural networ k built on a automatic differentiation system	& https://pytorch.org/ \\[0.5ex]
 \hline
CNTK	& A unified deep learning framework that uses a directed graph and series of computational steps to describe neural netowrk	& https://docs.microsoft.com/en-us/cognitive-toolkit/ \\[0.5ex]
 \hline
PyCaret	 & A machine learning framework for automation of machine learning workflows	& https://pycaret.org/ \\[0.5ex]
 \hline
DeepChem & A deep learning framework for Drug Discovery, Quantum Chemistry, Materials Science and Biology	 &https://deepchem.io/ \\[0.5ex]
 \hline
Deep Docking	& A deep learning framework for molecular docking	& https://github.com/zhenglz/dockingML \\[0.5ex]
 \hline
MolPAL	 & Active learning framework for high throughput virtual screeening	& https://github.com/coleygroup/molpal
\\[0.5ex]
 \hline
Hugging face	& A machine learning library that creates base model to build on top of tensorflow and Pytorch	& https://huggingface.co/ \\[0.5ex]
 \hline
GraphInvent	 & A platform to generate graph based molecules using GNN & https://github.com/MolecularAI/GraphINVENT \\[0.5ex]
 \hline
ATOMAI	& Deep learning framework for microscopy data	& https://atom.io/packages/ide-python \\[0.5ex]
 \hline
Veles	& A distributed DL framework	& 
 https://github.com/Samsung/veles
\\[0.5ex]
 \hline
 \end{longtable}

 \section{Important methods: }
 
The recent surge in Artificial Intelligence can be attributed to the availability of datasets and computing power such as Graphics Processing Units, Tensor Processing Units, and other hardware accelerators. These have allowed for the development of intricate and deep structures, which enable for exhaustive investigations such as Material Engineering. In this part, we will go over some of the AI methods used in Material Engineering. To begin, we will review the datasets that are available for AI based material engineering.

\subsection{Datasets for material engineering:}

Datasets are the fuel of any AI modal. For AI-based material engineering, we need datasets with desired properties and with a sufficient number of samples. Recently, the national institute of standards and technology (NIST) has produced several datasets for material engineering. The JARVIS-DFT is one such dataset.\cite{Choudhary_2020}  This dataset contains DFT-based material properties of 40000 bulk and 1000 crystalline materials. This dataset contains the properties such as formation energies, bandgaps, elastic, piezoelectric, dielectric constants, magnetic moments,  exfoliation energies for van der Waals bonded materials, improved meta-Generalized Gradient Approximations(meta-GGA) bandgaps, frequency-dependent dielectric function, spin-orbit spillage, spectroscopy limited maximum efficiency (SLME), infrared (IR) intensities, electric field gradient (EFG), heterojunction classifications, and Wannier tight-binding Hamiltonians.\cite{Choudhary_2020} The next dataset is the Open Quantum Material Database. This dataset is developed and maintained by the Wolverton Research Group at Northwestern University. This dataset contains the thermodynamic and structural properties of 1,022,603 materials.\cite{Saal2013} A detailed list of all the available datasets for AI-based material Engineering is provided in Table \ref{table3} of the supplementary document. 

\subsection{Neural network:}

\begin{figure}[!h]
 \centering
\includegraphics[width=.9\linewidth]{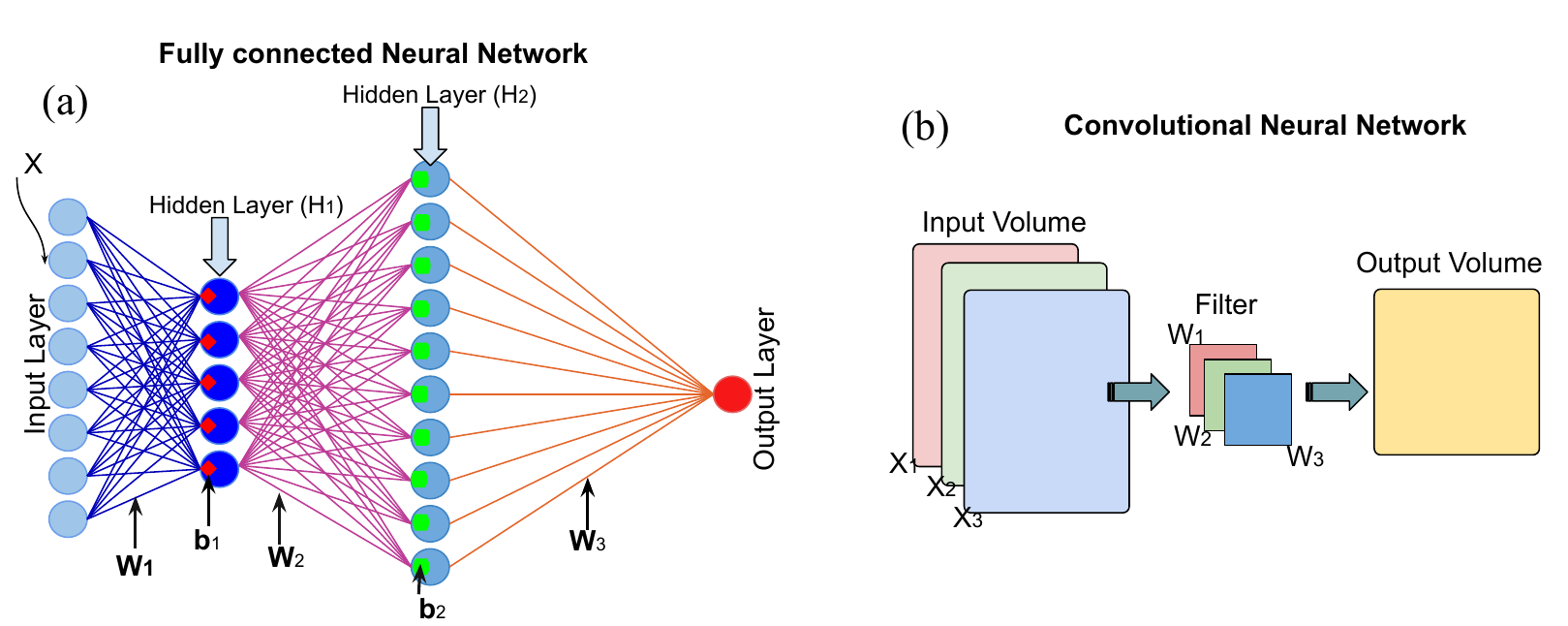}
\caption{A schematic of neural netowrks. (a) Schematio of fully connected neural network, (b) convolutional neural network}
\label{NN}
\end{figure}

Neural networks are modeled after the biological brain. A neural network consists of three types of layers. These are the input, hidden, and output layers. The input layer is the data feeding layer, the output layer is the final prediction layer and in the hidden layer(s) the feature extraction and computations are executed. Each layer in the network consists of a set of artificial neurons, which are connected to the neurons in the adjacent layers by a set of weights. The weights determine the strength of the connection between two neurons and are the parameters that the network learns during training. The basic operation of a neuron is to take a weighted sum of its inputs, apply an activation function to the result, and pass the output to the neurons in the next layer. The activation function is typically a non-linear function that introduces non-linearity into the network, allowing it to learn complex relationships in the data. A typical Nural Network is as shown in figure \ref{NN}. Mathematically the forward pass can be derived as 
$H_{i}=\sigma\left(X_{i}*W_{i}+b_{i}\right)$. Where $X_i$ is the input, $W_i$ is the weight matrix, $b_i$ is the bias and $\sigma$ is the activation function, such as sigmoid $\sigma(x)={\frac {1}{1+e^{-x}}}$. The learning process happens by adjusting the weights and biases to achieve the desired output at the output layer. This model can be further deepened by introducing more hidden layers. The deeper the layer the more complex phenomenon can be learned. This type of deep neural network can learn more complex features such as properties of materials. There are few other types of neural network that are extensively used in material engineering. These are Convolutional Neural Network (CNN), Graph Neural Networks (GNNs), Generative model etc. 

%===================

A convolutional neural network(CNN) is based on the mathematical operation of convolution, which is a way of combining two functions to produce a third function that represents how one of the original functions is modified by the other.\cite{CNN} In the case of a CNN, the input data (e.g., an image) is convolved with a set of learnable filters to produce a set of feature maps. In the context of a CNN, the input data is typically a 3D tensor with dimensions (width, height, channels), where channels represent the number of color channels (e.g., 3 for RGB images). The filters, or kernels, are also 3D tensors with dimensions (kernel width, kernel height, input channels), where input channels matches the number of channels in the input data.
To apply a convolutional layer in a neural network, slide the kernel over the input data, computing the dot product between the kernel and the local region of the input data at each position. This operation produces a single value, which is used to populate the corresponding position in the output feature map.
The mathematical formula for the convolution operation can be expressed in matrix form as follows:
 G= F * K
where F is a matrix representation of the input data, K is a matrix representation of the kernel, and G is a matrix representation of the output feature map.\cite{CNN} Zhuo Cao et al. have demonstrated the use of CNN in predicting the properties of crystalline materials.\cite{crystal}. However, performance of CNN generally suffers if the topology is arbitrary or change in orientation of the object. These type of limitations can be avoided using the graph neural network.\cite{77}

%==================
\subsection{Graph neural networks for efficient material engineering:} 

\begin{figure}[ht]
 \centering
\includegraphics[width=.8\linewidth]{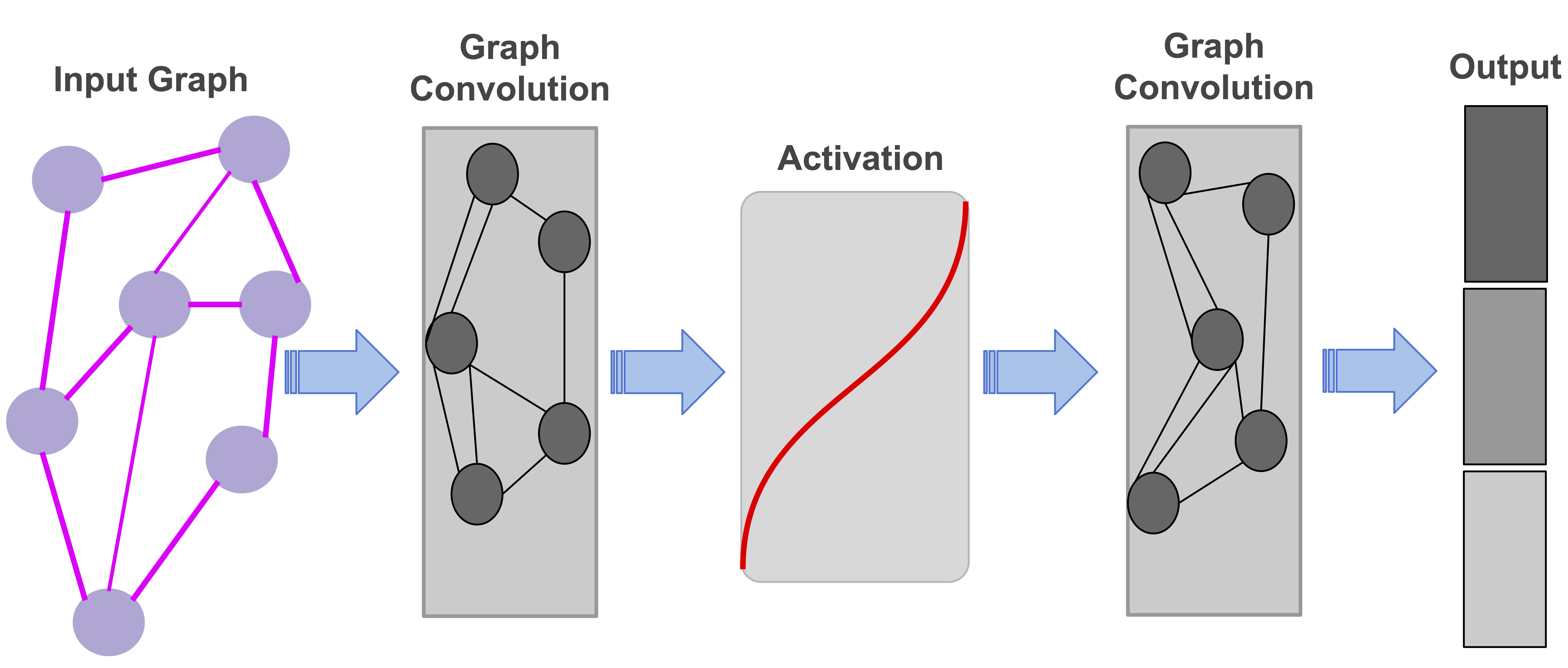}
 \caption{Schematic of a Graph Neural Network (GNN). Graph Neural Network is a type of Deep Learning Network used for node-level or edge-level predictions in graph data. This figure is used to show the mechanism of a GNN and its convolution process to produce embeddings of nodes and pass through the activation function to find out the output class}
 \label{fig5}
\end{figure}

Graph Neural Network (GNN) is an artificial neural network that processes and analyzes data structured in graph form. They have grown in popularity due to their capacity to deal with complex associations and dependencies between entities in a graph. Graphs consist of nodes (also known as vertices) and edges that link the nodes. Each node and edge can have connected characteristics, representing properties such as node or edge types, numerical values, or categorical labels. GNNs manipulate a graph by iteratively gathering information from neighboring nodes and edges, using a neural network to change and update the node representations. This process can be repeated over multiple layers, each learning more intricate patterns and dependencies. GNNs have produced positive results in many areas, for instance, recommendation systems, drug discovery, social network analysis, and traffic prediction. They can manage different graphs, such as directed and undirected, bipartite, and heterogeneous graphs.
A convenient way to represent molecules is in the form of a graph, where atoms are used to represent the featured nodes, and the interatomic bonds (with bond order) are used to represent edges. Features include properties like atomic identity, formal charge, and aromaticity.\cite{88} These are used in a molecular fingerprint that determines the presence or absence of a substructure in an atom. A typical graph neural network is shown in figure \ref{fig5}.

In GNN based model, interactions between atoms are learned by the atom embedding in a high-dimensional space and updating the embeddings by performing message passing. In recent years, many groundbreaking works have been published in GNN.\cite{12,15,44,53,54,60}. In \cite{28} crystal graph is used in property prediction. A similar architecture was applied to predict thermoelectric properties on a dataset of crystal and atomic information.\cite{12} The atomistic line graph neural network (ALIGNN) has achieved up to 85\% accuracy in solid and molecular property prediction.\cite{15}
In \cite{12}, the angular information is included explicitly by introducing a line graph. The line graph contains bond distance and bond angle information. The model designed with the inclusion of structural information is trained on crystalline materials properties and on molecular properties that could give good accuracy. This model outperformed conventional descriptors such as CFID.
Despite the great potentials of GNN, most of the state-of-the-art GNN models suffers from the issue of over-smoothing where with increasing depth, the model tends to make the embedding of all the nodes similar, which makes it challenging to classify unlabeled node and thus making GNN unscalable. To address this issue, an architecture called Deeper Graph Attention Neural Network (DeeperGATGNN) is introduced.\cite{96}. Before DeeperGATGNN, a simpler version named GATGNN was implemented. It contains two different soft attention layers; the first layer extracts the features at the local level of neighboring atoms. The subsequent attention layer extends this neighborhood-dependent information to the global context. This architecture is then extended by introducing additive skip connections between these soft attention layers and differentiable group normalization (DGN) layers. The DGN makes different clusters of the nodes of a graph, and each cluster is normalized separately, thus having different mean and standard deviation for different clusters and, thus, representation vector being dissimilar. Another advancement is the invocation of residual skip connection, where the stake layers learn the identity mapping denoted as $F(x) = H(x)- x$. Here x is the input, $ H(x)$ is the mapping function, and  $F(x)$ is the output function. This type of GNN claiming more accurate in predicting material property. 

%=================

\subsection{Generative models:}

\begin{figure}[!h]
 \centering
\includegraphics[width=.7\linewidth]{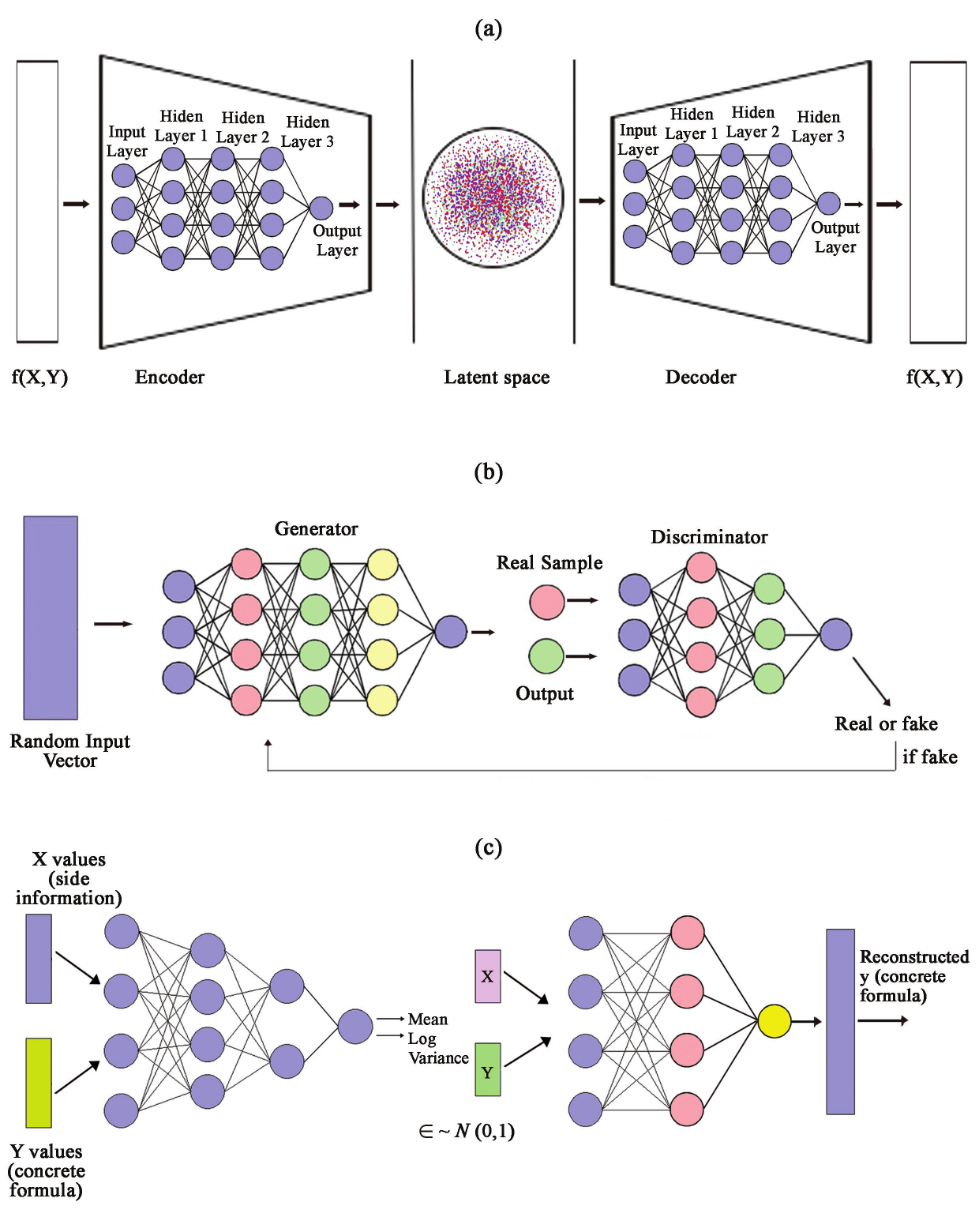}
\caption{Three Generative Models \textbf{a)}Schematic of the VAE model: Schematic of a Variational Autoencoder. The encoder acts as a compressor and generates the latent space by mapping the molecules to a vector space which is then mapped back to the molecule representation using the decoder.\cite{77} \newline \textbf{b)}Schematic of a Generative Adversarial Network. Two different convolutional neural networks are used in the model. One is used as a generator that generates some pattern from a set of the random input vector, and the discriminator network discriminates whether the data is real data or fakely generated by the model. If the data generated by the generator is labeled as fake by the discriminator, it is backpropagated to the generator. The generator readjusts the weight and resends the improved result to the discriminator. this process continues until it becomes difficult to differentiate between real data and the generator's data.\newline{c)The CVAE model is designed for new concrete formulas for building materials}}
\label{fig4}
\end{figure}

"One of the continuing scandals of physical science is that it remains, in general, impossible to predict the structure of even the simplest crystalline solids from a knowledge of their chemical composition." as quoted by John Maddox about 30 years ago.\cite{76} Over the years, though different methods are evolved for crystal structure prediction of molecules and solids, they are computationally expensive, and vast structure space is needed to be searched. As DL algorithms are introduced in this field, Generative Networks(GN), a class of DL algorithms, can tackle this problem to a great extent. 
 GN has the capability of producing samples from a given distribution. The functionality is given as input, and the model outputs a distribution of possible structures. This network works on joint probability distribution p(x, y), which means they can notice both the molecular representation of a material denoted as x and its property as y. If the conditional probability is applied, then the notation for design will be $(p(y|x))$, and by reversing the notation, we get $((p(x|y))$, which represents the inverse design of the material.\cite{77} 
The two most commonly used generative networks are:  Variational Autoencoder (VAE) and Generative Adversarial Network (GAN) (see figure \ref{fig4}).

The GAN is extended by adding an extra condition in both generator and discriminator by feeding an input layer Y to both networks. This extension is Conditional GAN(CGAN).\cite{100}. In an attempt to generate chemical compositions of inorganic materials without the crystal structure information, the concept of CGAN is used by Sawada et al. \cite{63}. In this model, the popular feature engineering scheme Bag-of-Atoms is used with the additional inclusion of the physical descriptors. In Bag-of-Atoms, the number of atoms in a chemical composition is represented in vector form. The physical descriptors are the fundamental properties of atoms that do not contain crystal information. CGAN is also used in \cite{83} for generating new concrete formulas for building materials. In the conditional probability $p(x|y)$, x contains the information about strength, age, and environmental impact, and y represents the number of constituent materials.

%\break

%========== Transfer of Learning==========

\subsection{Leveraging transfer learning for improved accuracy:} Needless to say that DL models are data hungry and can give plausible accuracy only after training with millions of data. But experimental datasets are normally smaller compared to computational datasets. The concept of deep transfer learning is therefore embedded in many of the recently developed DL models \cite{9}\cite{21},\cite{29}\cite{73}\cite{74}  where knowledge of one model is transferred to other model by making use of the features learned from the huge dataset. To accomplish this, at first,a source model is trained using a large dataset and then model parameters are tweaked by training on target property.
\begin{figure}[ht]
 \centering
\includegraphics[width=0.6\textwidth]{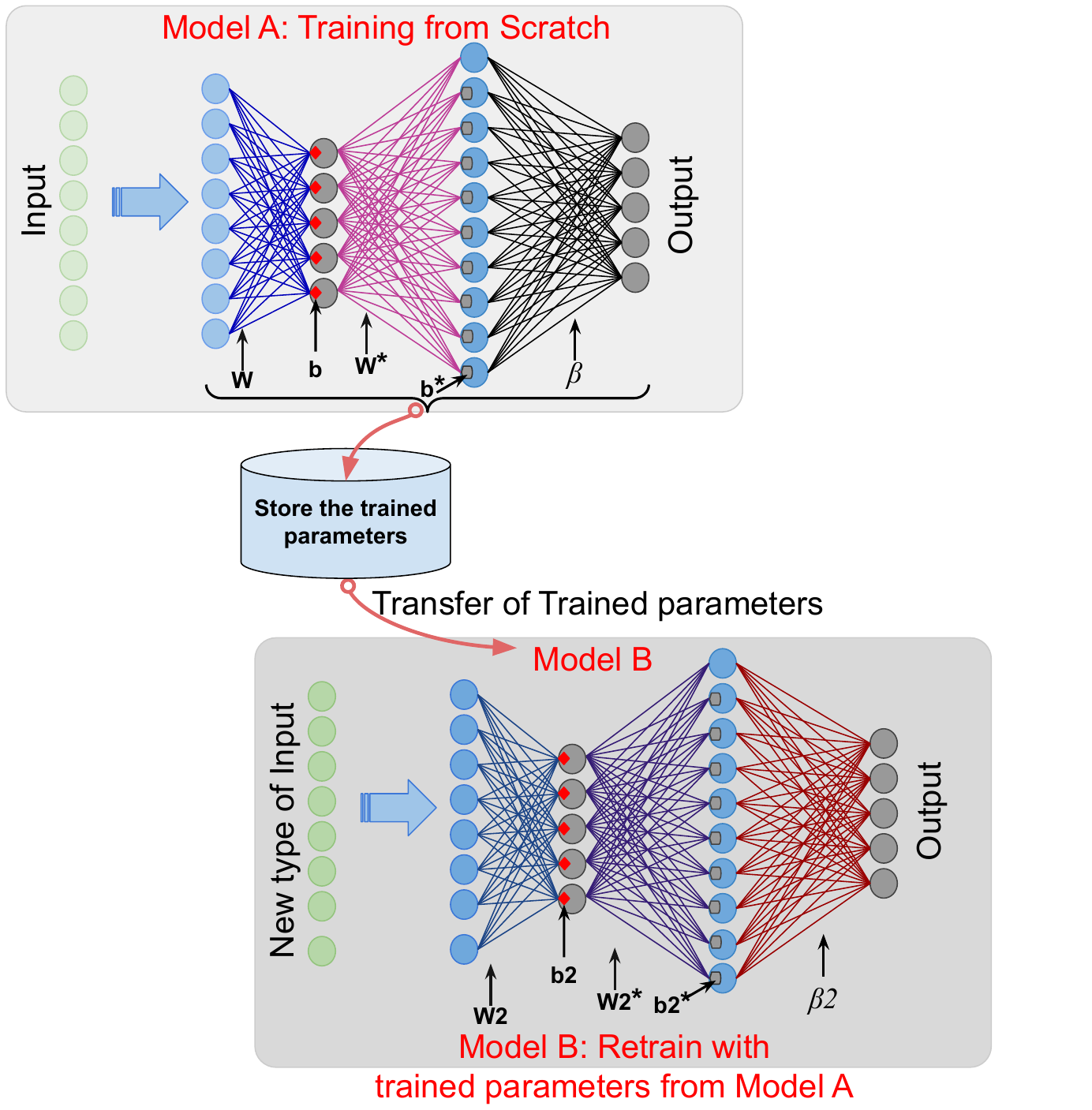}
 \caption{ Schematic representation of transfer learning. The DL model A is trained from scratch, especially with a large dataset. The model parameters like weights (W) and biases (b) are then saved. These saved parameters can be used whenever the model needs to be trained with a new type of dataset or any change needed in the architecture itself. In this scenario, the saved parameters can be transferred to model B and re-train with a new type of dataset. This method generally converges early compared to training the model from scratch.}
\label{fig3}
    
\end{figure}

A deeper concept of this is the cross-property deep transfer learning (see figure \ref{fig3}), where the source model and target models are trained for different properties.\cite{11}. The properties that don't have a large dataset can be predicted using this concept. It is implemented by two methods: i) by fine tuning the source model with target data and ii) learning the features from target dataset and then leveraging those in building target model.
The Original ElemNet DL model has gone through some significant improvements before implementation of cross property deep transfer learning which results in improved Mean Absolute Error (MAE). The modified version of ElemNet is implemented using TensorFlow 2 (TF2) with Keras as an interface instead of TensorFlow 1. These improvements in the library lead the model to learn chemical interactions and elemental similarities more accurately resulting a reduced in MAE of 0.0405 eV/atom. Again, the use of Monte Carlo Drop out in training, valiation and test phases leads to different activation and output for the same input in each run of the model which is a barrier in consistent feature extraction. The model improved its MAE further from 0.0405 eV/atom to 0.0373 eV/atom once this drop out is disabled. These to modifications leads to approx 10\% improvement in MAE.

%============

\subsection{AI techniques used in traditional analytical instruments: }
AI techniques are used extensively and have shown great potential in the existing traditional analytical instruments of MSE, such as X-Ray Diffraction (XRD), TEM, and Scanning Electron Microscopy (SEM) image analysis. In XRD, traditional descriptors that are used for pattern analyzation and for mapping the phase diagrams, are found to be tedious and time-consuming. Recent works made a leap forward through the seamless integration of ML models in depicting compounds of interest and phase diagrams. In \cite{90}, a Convolutional Neural Network (CNN) is used for one-to-one identification of XRD spectra of materials processed from an experimental database by removing noise. While comparing the performance of CNN with some other classical ML algorithms, it has shown better results with an accuracy of 96.7\%. 
In \cite{92}, a convolutional network and a dense network are used to learn the features of the inference patterns of powdered XRD, which will lead the model to predict the crystal symmetry, which was otherwise done conventionally using some peak finding algorithm. The convolutional network performed well in theoretical data, but it showed poor performance in experimental data, whereas the dense network showed higher classification accuracy for the experimental data. 82\% classification accuracy is achieved for the dense network while classifying only half of the samples. At the same time, the CNN could not give good accuracy to be used as in crystal symmetry prediction.
 In Transmission Electron Microscopy (TEM) also, AI has shown its potential. An unsupervised ML algorithm (Auto detect mNP) has been designed for the classification of particle shapes of metal nanoparticles (mNP) from TEM images. The algorithm can also determine the impurities in mNP synthesis. The algorithm is also able to classify the long rod and short rod nano particles.\cite{93}
For automation of structural analysis of nanoparticles from high-resolution TEM (HRTEM) data, a two-stage framework is designed. A CNN with a U-net architecture performs segmentation on HRTEM nanoparticles, and a random forest classifier detects the defects of individual nanoparticle regions with an accuracy of 86\%.\cite{52}

%=========
\subsection{ Advantage and disadvantages of AI in material engineering and its future:}

The use of AI in material engineering is a dynamic area of research that is evolving continuously. Using the right kind of algorithm can significantly accelerate the discovery of new materials as ML algorithms can learn repetitive patterns, resulting in faster simulation of complex structures and chemical reactions.
The challenge of building interatomic potentials is also taken care of by AI-based methods, as the artificial Neural Network has paved the way for the construction of potential energy surfaces with a higher efficiency of several orders of magnitude compared to traditional methods.
AI algorithms can also be used for exploring the massive chemical space of a material by training the model with existing samples and then using the trained model for predicting all possible combinations, which was earlier a big hurdle for simulation-based methods such as DFT. Thousands of stable configurations have been discovered by researchers by training ML models with the existing datasets. Future researchers can work for more efficient feature space extraction to provide new paradigms for discovering stable material configurations.
ML has stepped into the area of drug discovery too. The Graph Neural Networks and other DL models have greatly helped researchers predict the solubility of molecules and drug target interactions for drug material. Another critical area of research in Drug discovery is the prediction of the protein structure of the targeted molecule to make the treatment successful. Here the AI researcher can grab the opportunity and think big for developing ML and DL algorithms to fuel the growth of target protein structure prediction.
 In energy storage materials, algorithms can be developed to fulfill multiple objectives. For example, properties like dipole orientation, atomic polarization, resonant effect, and relaxation effect are to be predicted simultaneously to screen a dielectric material. These factors may also have inverted coupling relationships to be optimized collaboratively. For optimizing collective properties, single optimization algorithms are not enough. Therefore, in future work, people may go for the development of multi-optimization algorithms.
 Though ML has revolutionized the discoveries of material science, there are a few challenges that still need to be addressed. 
 The lack of data (especially from experimental results) is one of the significant drawbacks. Again the poor explainability of AI models is another drawback. The complex generative models are generally treated as a black box, where chemical relationships are not firmly established, and the error analysis too is a difficult task. Therefore, the development of both highly accurate and interpretable models will definitely make a leap forward in material science research.
 The non-viability is a significant hurdle faced by the chemical structures generated by the generative deep learning models. The highly complex chemical structures produced by the generative models outside the existing chemical space are theoretically feasible. However, synthesizing them may not be viable because of the high cost and complexity.
%=========

\section{Conclusion:}
In conclusion, AI will radically change the ways of developing materials. This will help us to find the materials of desired properties efficiently. This will reduce the time, money, and effort that are needed to create a material. The capabilities of traditional analytical tools will also be improved by adopting the advantages of AI and ML. Recently applied graph neural network-based approaches and generative adversarial algorithms are helping in the rapid development of drugs. The Google Alpha Fold is one such example, which helps us to find the folding of proteins with very high accuracy and thereby find appropriate molecules for drugs. Again various computer vision-based tools are helping us to analyze traditional analytical tools such as transmission electron microscopy (TEM),  scanning electron microscope (SEM), Atomic Force Microscopy (AFM), etc. The traditional computation tools such as DFT and in-slico methods are also defeated by AI and ML-based algorithms, both in terms of accuracy and efficiency. All these development have brought us the hope to engineer materials that can meet the demands and also shape the future.

\section*{Acknowledgment}
Mohendra Roy acknowledges the seed grant No. $ORSP/R\&D/PDPU/2019/MR/RO051$ of PDEU, the core research grant No. $CRG/2020/000869$ of the Science and Engineering Research Board (SERB), Govt. of India and the project grant no $GUJCOST/STI/202-22/3873$ of GUJCOST, Govt. of Gujarat, India. \newline \\

\textcolor{red}{\textbf{Note:} The supplementary is available at the end of the reference section}

%\newline
\bibliographystyle{unsrt}  
\bibliography{Reference.bib}

\pagebreak

%\begin{appendices}

 \section*{Supplementary Tables}
 
\setlength{\tabcolsep}{.5pt} %% default is 6pt
\setlength\extrarowheight{0.5pt}
\begin{longtable}{|M{2.4cm}|M{2.8cm}|M{3.2cm}|M{2.5cm}|M{3cm}|M{3cm}|}
\caption{The first column of the table contains some of the latest AI, ML and DL models used for materials property prediction,second column indicating the type of property being predicted, third column gives the details of the predicted property, fourth column gives the performance of the respective models in different parameters and the final column categorises the model based on the techniques of that model. \label{table2}}\\
 \hline
 \multicolumn{6}{| c |}{Begin of Table}\\
 \hline
 Model Type & Property Type & Properties to Predict & \multicolumn{2}{c|}{Accuracy/Performance} & Models based on Techniques\tabularnewline
\hline 
\hline
 \endfirsthead
\hline
 \multicolumn{6}{|c|}{Continuation of Table \ref{table2}}\\
 \hline 
\hline 
\hline
\endhead

\hline
\endfoot
\hline
\multicolumn{6}{| c |}{End of Table}\\
\hline\hline
\endlastfoot
\hline 
 Active Learning + Roost+ MatGNN \ \cite{48} & Inverse Design & To discover new stable Materials and Semiconductor &  & discovered one high bandgap material and six semiconductor in a specified
range & Generative model + Generative adversarial network\tabularnewline
\hline 
\hline 
 \multirow{31}{*}{ALIGNN \cite{24}} & \multirow{31}{*}{Electronic} & Formation Energy (MP dataset) & \multirow{31}{*}{MAE} & 0.022ev/atom & \multirow{31}{*}{\makecell{Graph Neural \\ Network}}\tabularnewline
\cline{3-3} \cline{5-5} 
 &  & Bandgap(MP dataset) &  & 0.218 ev & \tabularnewline
\cline{3-3} \cline{5-5} 
 &  & Formation Energy(JARVIS-DFT dataset) &  & 0.033ev/at & \tabularnewline
\cline{3-3} \cline{5-5} 
 &  & Bandgap(JARVIS-DFT dataset) &  & 0.14 ev & \tabularnewline
\cline{3-3} \cline{5-5} 
  &  & Total energy  &  & 0.037 ev/at  & \tabularnewline
\cline{3-3} \cline{5-5} 
 &  & Ehull  &  & 0.076 ev   & \tabularnewline
\cline{3-3} \cline{5-5} 
 &  & Bandgap (MBJ)  &  & 0.31 ev     & \tabularnewline
\cline{3-3} \cline{5-5} 
 &  & Voigt bulk &  & 10.40 GPa     & \tabularnewline
\cline{3-3} \cline{5-5} 
 &  &  shear modulus &  & 9.48 Gpa    & \tabularnewline
\cline{3-3} \cline{5-5} 
 &  & Magnetic Moment &  & 0.26 \textgreek{a}|B  & \tabularnewline
\cline{3-3} \cline{5-5} 
 &  & Spectroscopic Limited Maximum Efficiency  &  & 4.52 No unit & \tabularnewline
\cline{3-3} \cline{5-5} 
 &  & Spillage &  & 0.35 No unit & \tabularnewline
\cline{3-3} \cline{5-5} 
 &  & Kpoint-length  &  & 9.51Å & \tabularnewline
\cline{3-3} \cline{5-5} 
 &  & Plane-wave cutoff &  & 133.8 eV & \tabularnewline
\cline{3-3} \cline{5-5} 
 &  & \ensuremath{\epsilon}x(OPT)   &  & 20.40 No unit & \tabularnewline
\cline{3-3} \cline{5-5} 
 &  & \ensuremath{\epsilon}y (OPT)  &  & 19.99 No unit & \tabularnewline
\cline{3-3} \cline{5-5} 
 &  & \ensuremath{\epsilon}z(OPT) &  & 19.57 No unit & \tabularnewline
\cline{3-3} \cline{5-5} 
 &  & \ensuremath{\epsilon}x (MBJ) &  & 24.05 No unit & \tabularnewline
\cline{3-3} \cline{5-5} 
 &  & \ensuremath{\epsilon}y (MBJ)  &  & 23.65 No unit & \tabularnewline
\cline{3-3} \cline{5-5} 
 &  & \ensuremath{\epsilon}z (MBJ)  &  & 23.73 No unit & \tabularnewline
\cline{3-3} \cline{5-5} 
 &  & \ensuremath{\epsilon}(DFPT:elec+ionic) &  & 28.15 No unit   & \tabularnewline
\cline{3-3} \cline{5-5} 
 &  & Max. piezoelectric strain coeff (dij)  &  & 20.57 CN\textminus 1  & \tabularnewline
\cline{3-3} \cline{5-5} 
 &  & Max. piezo. stress coeff (eij)  &  & 0.147 Cm\textminus 2 & \tabularnewline
\cline{3-3} \cline{5-5} 
 &  & Exfoliation Energy  &  & 51.42  mev(atom)-1 & \tabularnewline
\cline{3-3} \cline{5-5} 
 &  & Max. EFG &  & 19.12 1021 Vm\textminus 2 & \tabularnewline
\cline{3-3} \cline{5-5} 
 &  & avg. me  &  & 0.085 electron mass unit & \tabularnewline
\cline{3-3} \cline{5-5} 
 &  & avg.mh &  & 0.124 electron mass unit & \tabularnewline
\cline{3-3} \cline{5-5} 
 &  & n-Seebeck &  & 40.92\textgreek{m}VK\textminus 1 & \tabularnewline
\cline{3-3} \cline{5-5} 
 &  & n-PF &  & 442.3\textgreek{m}W(mK2)\textminus 1 & \tabularnewline
\cline{3-3} \cline{5-5} 
 &  & P-Seebeck &  & 42.42\textgreek{m}VK\textminus 1 & \tabularnewline
\cline{3-3} \cline{5-5} 
 &  & p-PF &  & 440.26\textgreek{m}W(mK2)\textminus 1 & \tabularnewline
 \hline
\multirow{9}{*}{CGCNN\cite{1}} & \multirow{9}{*}{Electronic/ Elastic} & Formation energy  & \multirow{7}{*}{MAE} & 0.039 ev/atom & \multirow{9}{*}{GNN}\tabularnewline
\cline{3-3} \cline{5-5} 
 &  &  Absolute Energy   &  &  0.072 ev/atom & \tabularnewline
\cline{3-3} \cline{5-5} 
 &  & Bandgap  &  & 0.388 ev & \tabularnewline
\cline{3-3} \cline{5-5} 
 &  & Fermi Energy  &  & 0.363 ev  & \tabularnewline
\cline{3-3} \cline{5-5} 
 &  & Bulk moduli   &  & 0.054 log(Gpa) & \tabularnewline
\cline{3-3} \cline{5-5} 
 &  & Shear moduli  &  & 0.087 log(Gpa) & \tabularnewline
\cline{3-3} \cline{5-5} 
 &  & Poisson ratio  &  & 0.030 log(Gpa) & \tabularnewline
\cline{3-5} \cline{4-5} \cline{5-5} 
 &  & classification of metal  & \multirow{2}{*}{for threshold 0.5} & 0.8 & \tabularnewline
\cline{3-3} \cline{5-5} 
 &  & classification of semiconductor  &  & 0.95 & \tabularnewline
\hline 
\multirow{9}{*}{\makecell{Conditional \\ Variational \\ Autoencoder \\ (CVAE) \cite{83}}} & \multirow{9}{*}{\makecell{To design \\ concrete formulas}} & \multirow{3}{*}{\makecell{Global warming \\ Potential (GWP) \\ (of the formulas)}} & MAE  & 7.187  & \multirow{9}{*}{Generative Model}\tabularnewline
\cline{4-5} \cline{5-5} 
 &  &  & RMSE  & 9.374  & \tabularnewline
\cline{4-5} \cline{5-5} 
 &  &  & R2  & 0.979  & \tabularnewline
\cline{3-5} \cline{4-5} \cline{5-5} 
 &  & \multirow{3}{*}{\makecell{Acidification \\ Potential (AP)\\ (of the formulas)}} & MAE  & 0.019  & \tabularnewline
\cline{4-5} \cline{5-5} 
 &  &  & RMSE  & 0.04  & \tabularnewline
\cline{4-5} \cline{5-5} 
 &  &  & R2  & 0.974  & \tabularnewline
\cline{3-5} \cline{4-5} \cline{5-5} 
 &  & \multirow{3}{*}{\makecell{Strength Predictor \\ Performance \\ (of the formulas)}} & MAE & 4.457 (>=90 days)  & \tabularnewline
\cline{4-5} \cline{5-5} 
 &  &  & RMSE & 0.125 (>=90 days)  & \tabularnewline
\cline{4-5} \cline{5-5} 
  &  &  & R2  & 0.789 (>=90 days)  & \tabularnewline
\hline
\multirow{4}{*}{\makecell{CVAE + \\ 3-D U-Net \\ segmentation \\ model \cite{4}}} & \multirow{4}{*}{\makecell{Encode and Decode \\ 3-D atomic position \\ and species}} & Segment the locations of molecules & \multirow{3}{*}{\makecell{for Single \\ Unit cell}} & 99\% & \multirow{4}{*}{\makecell{Generative Model + \\ CNN}} \tabularnewline
\cline{3-3} \cline{5-5} 
 &  &  Reconstruction  &  & 90\% & \tabularnewline
\cline{3-3} \cline{5-5} 
 &  &  Classification of Species &  & 66\% & \tabularnewline
\cline{3-5} \cline{4-5} \cline{5-5} 
 &  & Nearest Atom Species Prediction & for Repeating Unit cell & 65.40\% & \tabularnewline
\hline 
 \multirow{20}{*}{\makecell{Compositionally\\Restricted \\ Attention-based \\ Network \\(CrabNet) \cite{104}}} &  \multirow{20}{*}{\makecell{Electronic \\Thermal \\ Elastic}} & Castelli perovskites & \multirow{20}{*}{MAE} & 0.127 & \multirow{20}{*}{\makecell{Graph Attention \\ Network (GAN)}}\tabularnewline
\cline{3-3} \cline{5-5} 
 &  &  Refractive Index &  & 0.348 & \tabularnewline
\cline{3-3} \cline{5-5} 
  &  & Shear modulus &  & 0.092 & \tabularnewline
\cline{3-3} \cline{5-5} 
 &  & Bulk Modulus &  & 0.068 & \tabularnewline
\cline{3-3} \cline{5-5} 
 &  & Experimental band gap  &  & 0.338 & \tabularnewline
\cline{3-3} \cline{5-5} 
 &  & Exfoliation Energy &  & 50.512 & \tabularnewline
\cline{3-3} \cline{5-5} 
 &  & MP Formation Energy &  & 0.077 & \tabularnewline
\cline{3-3} \cline{5-5} 
 &  & MP Band gap &  & 0.263 & \tabularnewline
\cline{3-3} \cline{5-5} 
 &  & Phonon peak &  & 53.341 & \tabularnewline
\cline{3-3} \cline{5-5} 
  &  &   Steels yield &  & 91.748 & \tabularnewline
\cline{3-3} \cline{5-5} 
 &  & AFLOW Bulk Modulus &  & 8.692 & \tabularnewline
\cline{3-3} \cline{5-5} 
 &  & AFLOW Debye Temperature &  & 33.464 & \tabularnewline
\cline{3-3} \cline{5-5} 
 &  & ,AFLOW Shear Modulus, &  & 9.082 & \tabularnewline
\cline{3-3} \cline{5-5} 
 &  & AFLOW Thermal Conductivity, &  & 2.318 & \tabularnewline
\cline{3-3} \cline{5-5} 
 &  & AFLOW Thermal Expansion &  & 3.85E-06 & \tabularnewline
\cline{3-3} \cline{5-5} 
 &  & AFLOW Band gap &  & 0.301 & \tabularnewline
\cline{3-3} \cline{5-5} 
 &  & AFLOW Energy per atom, &  & 0.093 & \tabularnewline
\cline{3-3} \cline{5-5} 
 &  & Bartel Decomposition &  & 0.063 & \tabularnewline
\cline{3-3} \cline{5-5} 
 &  &  Bartel Formation,                                    &  & 0.059 & \tabularnewline
\cline{3-3} \cline{5-5} 
 &  & MP bulk modulus &  & 11.209 & \tabularnewline
\cline{3-3} \cline{5-5} 
 &  & MP Elastic anisotropy, &  & 8.263 & \tabularnewline
\cline{3-3} \cline{5-5} 
 &  & MP Energy above convex hull,  &  & 0.089 & \tabularnewline
\cline{3-3} \cline{5-5} 
 &  & MP Magnetic Moment &  & 2.105 & \tabularnewline
\cline{3-3} \cline{5-5} 
 &  &  MP shear modulus,                                   &  & 12.787 & \tabularnewline
\cline{3-3} \cline{5-5} 
 &  & OQMD Band gap &  & 0.049 & \tabularnewline
\cline{3-3} \cline{5-5} 
 &  & OQMD Energy per Atom,  &  & 0.033 & \tabularnewline
\cline{3-3} \cline{5-5} 
 &  & OQMD Formation Enthalpy &  & 0.031 & \tabularnewline
\cline{3-3} \cline{5-5} 
 &  &   OQMD Volume per atom &  & 0.277 & \tabularnewline
 \hline
 \multirow{6}{*}{\makecell{Crystal \\ eXplainable \\ Property  \\ Predictor\\ (CrysXPP)\cite{65}}} & \multirow{6}{*}{Crystal state,Elastic} & Formation energy & \multirow{6}{*}{MAE} & 0.086 ev/atom & \multirow{6}{*}{GNN}\tabularnewline
\cline{3-3} \cline{5-5} 
 &  & Band Gap  &  & 0.467 ev & \tabularnewline
\cline{3-3} \cline{5-5} 
 &  & Fermi Energy  &  &  0.471 ev & \tabularnewline
\cline{3-3} \cline{5-5} 
 &  & Bulk moduli  &  & 0.08 log(Gpa) & \tabularnewline
\cline{3-3} \cline{5-5} 
 &  & Shear moduli  &  & 0.105 log(Gpa) & \tabularnewline
\cline{3-3} \cline{5-5} 
 &  & Poisson ratio &  & 0.035 log(Gpa) & \tabularnewline
\hline 
\multirow{3}{*}{\makecell{Deep Adaptive \\ Regressive \\ Weighted \\ Intelligent \\ Network \\ (DARWIN) \cite{50}}} & \multirow{3}{*}{Electronic} & Band Gap (for Zn based systems) & \multirow{3}{*}{MAE} & 0.9592ev & \multirow{3}{*}{\makecell{Graph Convolutional \\ Neural Network \\ (GCNN)}}\tabularnewline
\cline{3-3} \cline{5-5} 
 &  & Band Gap (for Cu based systems) &  & 0.1903ev & \tabularnewline
\cline{3-3} \cline{5-5} 
 &  & Band Gap (for Mg based systems) &  & 1.4995ev & \tabularnewline
\hline
\multirow{6}{*}{\makecell{DeeperGATGNN \\ \cite{60}}} & \multirow{6}{*}{Physico-Chemical} & Bulk Materials Formation Energy & \multirow{6}{*}{\makecell{over previous \\ best model}} & 34.97\% ev/atom & \multirow{6}{*}{GNN}\tabularnewline
\cline{3-3} \cline{5-5} 
 &  &  Alloy Surface Adsorption Energy &  & 29.55\% & \tabularnewline
\cline{3-3} \cline{5-5} 
 &  &  Pt-cluster Total Energy &  & 14.03\% & \tabularnewline
\cline{3-3} \cline{5-5} 
 &  & 2D Materials Work Function &  & 15.76\% & \tabularnewline
\cline{3-3} \cline{5-5} 
 &  & MOF Band Gap  &  & 5.34\% & \tabularnewline
\cline{3-3} \cline{5-5} 
 &  & Bulk Materials Bandgap  &  & 5.42\% & \tabularnewline
\hline 
\multirow{12}{*}{\makecell{\\ (DimeNet++) \\\cite{95}}} & \multirow{12}{*}{\makecell{Thermodynamic, \\ Electronic}} & Dipole Moment (mu) & \multirow{12}{*}{MAE} & 0.0297 D & \multirow{12}{*}{GNN}\tabularnewline
\cline{3-3} \cline{5-5} 
 &  & Electronic Polarizability (alpha) &  & 0.0435a03 & \tabularnewline
\cline{3-3} \cline{5-5} 
 &  &  HOMO &  & 24.6meV & \tabularnewline
\cline{3-3} \cline{5-5} 
 &  & LUMO &  & 19.5 & \tabularnewline
\cline{3-3} \cline{5-5} 
 &  &  Energy difference of HOMO and LUMO &  & 32.6meV & \tabularnewline
\cline{3-3} \cline{5-5} 
 &  & electronic spatial extent <R2> &  & 0.331a02 & \tabularnewline
\cline{3-3} \cline{5-5} 
 &  &  ZPVE &  & 1.21meV & \tabularnewline
\cline{3-3} \cline{5-5} 
 &  & Internal Energy at 0k (U0), &  & 6.32meV & \tabularnewline
\cline{3-3} \cline{5-5} 
 &  & Internal Energy at 298K (U) &  & 6.28meV & \tabularnewline
\cline{3-3} \cline{5-5} 
 &  & enthalpy at 298 K(H) &  & 6.53meV & \tabularnewline
\cline{3-3} \cline{5-5} 
 &  & Gibbs free energy at 298 K(G), &  & 7.56meV & \tabularnewline
\cline{3-3} \cline{5-5} 
 &  & heat capacity at 298 K (Cv) &  & 0.0230cal/mol K & \tabularnewline
 \hline
 
Decision-Trees (DT) \cite{10} & \multirow{7}{*}{Classification Task} & \multirow{7}{*}{\makecell{Classify \\ materials \\ based on \\ spectroscopic \\ limited maximum \\ efficiency
(SLME)}} & \multirow{7}{*}{AUC} & 0.67 & \multirow{7}{*}{Classification Model}\tabularnewline
\cline{1-1} \cline{5-5} 
randomforest (RF) \cite{10} &  &  &  & 0.79 & \tabularnewline
\cline{1-1} \cline{5-5} 
K-nearest neighbor (KNN) \cite{10} &  &  &  & 0.77 & \tabularnewline
\cline{1-1} \cline{5-5} 
Multi-layer perceptron (MLP) \cite{10} &  &  &  & 0.8 & \tabularnewline
\cline{1-1} \cline{5-5} 
GBDT in scikit learn (SK-GB)\cite{10} &  &  &  & 0.84 & \tabularnewline
\cline{1-1} \cline{5-5} 
GBDT in XGBoost (XGB) \cite{10} &  &  &  & 0.84 & \tabularnewline
\cline{1-1} \cline{5-5} 
GBDT in LightGBM (LGB) \cite{10} &  &  &  & 0.87 & \tabularnewline
\hline 
ElemNet \cite{109} & Electronic & Formation Enthalpy & MAE & 0.050\textpm 0.0007 ev/atom & --------- \tabularnewline
\hline 
Fully Connected Neural Network (FCNN)\cite{111} & Thermoelectric Property & Thermoelectric power factor & MAE & 20.70\% & FCNN\tabularnewline
\hline 
\multirow{4}{*}{\makecell{Heuristic and \\ Ensemble of \\ decision trees\\ \cite{107}}} & \multirow{4}{*}{Electronic} & \multirow{4}{*}{Formation Energy} & MAE(Decision Trees)                & 0.16ev  & \multirow{4}{*}{ --------------- }\tabularnewline
\cline{4-5} \cline{5-5} 
 &  &  &  R2(Decision Trees) & 0.93 & \tabularnewline
\cline{4-5} \cline{5-5} 
 &  &  & MAE (heuristic) &  0.12ev & \tabularnewline
\cline{4-5} \cline{5-5} 
 &  &  & R2(Heuristic) & 0.95 & \tabularnewline
 \hline 
\hline 
\multirow{4}{*}{\makecell{Hierarchically \\ Interacting \\ Particle \\ Neural\\ Network \\ (HIP-NN) \cite{33}}} & \multirow{4}{*}{Quantum Chemical} & Energy of Benzene,  & \multirow{4}{*}{\makecell{MAE (for training \\ size=50k)}} & 0.064 \textpm{} 0.002    kcal/mol & \multirow{4}{*}{\makecell{Residual Neural \\ Network}}\tabularnewline
\cline{3-3} \cline{5-5} 
 &  & Energy of Malonaldehyde &  & 0.094 \textpm{} 0.001     kcal/mol & \tabularnewline
\cline{3-3} \cline{5-5} 
 &  & Energy of Salicylic Acid &  & 0.195 \textpm{} 0.002    kcal/mol & \tabularnewline
\cline{3-3} \cline{5-5} 
 &  & Energy of Toluene &  &  0.144 \textpm{} 0.004    kcal/mol & \tabularnewline
\hline 
\multirow{10}{*}{Hydra-GNN \cite{30}} & \multirow{10}{*}{Magnetic, Electronic} & \multirow{3}{*}{\makecell{Multi-task \\ Learning(MTL),\\ Mixing Enthalpy(H) \\ Charge transfer(C) \\Magnetic
Moment \\ (M) }} & \multirow{10}{*}{RMSE   } & H= 7.54e\textminus 3 \textpm{} 8.70e\textminus 4  & \multirow{10}{*}{Multi-Task GCN}\tabularnewline
\cline{5-5} 
 &  &  &  &  C= 6.77e\textminus 3 \textpm{} 3.59e\textminus 4 1 & \tabularnewline
\cline{5-5} 
 &  &  &  & M= 04e\textminus 2 \textpm{} 4.94e\textminus 4 & \tabularnewline
\cline{3-3} \cline{5-5} 
 &  & \multirow{2}{*}{MTL, HC} &  & H=7.33e-3 \textpm{} 4.77e-4 & \tabularnewline
\cline{5-5} 
 &  &  &  & C=7.36e-3 \textpm{} 3.23e-4 & \tabularnewline
\cline{3-3} \cline{5-5} 
 &  & \multirow{1}{*}{MTL, HM} &  & H=6.64e-3 \textpm{}  5.08e-4  & \tabularnewline
\cline{5-5} 
 &  &  &  & M=1.02e-2 \textpm{} 5:23e-4 & \tabularnewline
\cline{3-3} \cline{5-5} 
 &  & STL(Single-Task Learning) H &  & H=1.02e-2 \textpm{} 1.16e-3 & \tabularnewline
\cline{3-3} \cline{5-5} 
 &  & STL, M &  & M=8.77e-3\textpm{} 3.18e-4 & \tabularnewline
\cline{3-3} \cline{5-5} 
 &  & STL, C &  & C=5.94e-3 \textpm{} 4.39e-4 & \tabularnewline
\hline 
\multirow{10}{*}{IRNET \cite{103}} & \multirow{10}{*}{\makecell{Electronic and \\ Elemental}} & OQMD-C Formation Enthalpy  & \multirow{10}{*}{\makecell{MAE ev/atom \\ (for 17 \\ Layer IRNET)}} & 0.054 & \multirow{10}{*}{\makecell{Residual Neural\\ Network}}\tabularnewline
\cline{3-3} \cline{5-5} 
 &  &  OQMD-C Bandgap &  & 0.051 & \tabularnewline
\cline{3-3} \cline{5-5} 
 &  & OQMD-C Energy per atom  &  & 0.0696 & \tabularnewline
\cline{3-3} \cline{5-5} 
 &  & OQMD-C Volume per atom &  & 0.415 & \tabularnewline
\cline{3-3} \cline{5-5} 
 &  &   MP-C Bandgap &  & 0.363 & \tabularnewline
\cline{3-3} \cline{5-5} 
 &  & MP-C density &  & 0.348 & \tabularnewline
\cline{3-3} \cline{5-5} 
 &  & MP-C Enery-above-hull   &  & 0.091 & \tabularnewline
\cline{3-3} \cline{5-5} 
 &  & MP-C Enery-per-atom &  & 0.143 & \tabularnewline
\cline{3-3} \cline{5-5} 
 &  & MP-C Total magnetization &  & 3.005 & \tabularnewline
\cline{3-3} \cline{5-5} 
 &  &  MP-C Volume &  & 215.037 & \tabularnewline
\hline 
\hline 
JARVIS-STMnet \cite{8} & Classification Task &  Classification of five lattice classes (square, hexagon, rhombus/centered-rectangle,
rectangle and parallelogram/oblique) &  & 0.9 & Convolutional Neural Network (CNN)\tabularnewline
\hline 
MATGNN \cite{108} & Inverse design & Generates hypothetical inorganic materials &  & 84.5\%  chemically valid samples out of total generated samples & Generative adversarial network\tabularnewline
\hline 
Message Passing Neural Network (MPNN)(enn-s2s) \cite{102} & \multirow{2}{*}{Chemical Property} & \multirow{2}{*}{\makecell{mu,alpha,HOMO,\\LUMO,Gap,R2,\\ZPVE,U0,U,H,\\ G, Cv,Omega,}} & \multirow{2}{*}{MAE(avg)} & 0.68 & \multirow{2}{*}{GNN}\tabularnewline
\cline{1-1} \cline{5-5} 
MPNN (enn-s2s-ens5) \cite{102} &  &  &  & 0.52 & \tabularnewline
\hline 
\multirow{17}{*}{\makecell {Moleculenet \cite{36}}} & \multirow{17}{*}{\makecell { Quantum mechanical,\\ physical chemistry,\\ biophysical affinity ,\\ physiological} } & Properties on QM7 dataset             & MAE (on best model) & DTNN: 8.75 & \multirow{17}{*}{\makecell{Different Variations \\ of GNN}}\tabularnewline
\cline{3-5} \cline{4-5} \cline{5-5} 
 &  & Properties on QM7b dataset & MAE & DTNN: 1.77a & \tabularnewline
\cline{3-5} \cline{4-5} \cline{5-5} 
 &  & Properties on QM8 dataset  & MAE & MPNN: 0.0143 & \tabularnewline
\cline{3-5} \cline{4-5} \cline{5-5} 
 &  & Properties on QM9 dataset  & MAE &  DTNN: 2.35 & \tabularnewline
\cline{3-5} \cline{4-5} \cline{5-5} 
 &  & Properties on ESOL    & RMSE    & MPNN: 0.58 & \tabularnewline
\cline{3-5} \cline{4-5} \cline{5-5} 
 &  & Properties on FreeSolv & RMSE & MPNN: 1.15 & \tabularnewline
\cline{3-5} \cline{4-5} \cline{5-5} 
 &  & Properties on Lipophilicity      & RMSE & GC: 0.655 & \tabularnewline
\cline{3-5} \cline{4-5} \cline{5-5} 
 &  & Properties on PCBA & AUC-PRC & GC: 0.136 & \tabularnewline
\cline{3-5} \cline{4-5} \cline{5-5} 
 &  &Properties on  MUV    & AUC-PRC & Weave: 0.109 & \tabularnewline
\cline{3-5} \cline{4-5} \cline{5-5} 
 &  & Properties on HIV & AUC-PRC & GC: 0.763  & \tabularnewline
\cline{3-5} \cline{4-5} \cline{5-5} 
 &  & Properties on BACE  & AUC-PRC & Weave: 0.806 & \tabularnewline
\cline{3-5} \cline{4-5} \cline{5-5} 
 &  & Properties on PDBBind(FULL)  & RMSE & GC: 1.44 & \tabularnewline
\cline{3-5} \cline{4-5} \cline{5-5} 
 &  & Properties on BBBP & AUC-PRC & GC: 0.690  & \tabularnewline
\cline{3-5} \cline{4-5} \cline{5-5} 
 &  & Properties on Tox21 & AUC-PRC & GC: 0.829 & \tabularnewline
\cline{3-5} \cline{4-5} \cline{5-5} 
 &  & Properties on ToxCast  & AUC-PRC & Weave: 0.742 & \tabularnewline
\cline{3-5} \cline{4-5} \cline{5-5} 
 &  & Properties on SIDER  & AUC-PRC & GC: 0.638 & \tabularnewline
\cline{3-5} \cline{4-5} \cline{5-5} 
 &  & Properties on ClinTox & AUC-PRC & Weave: 0.832 & \tabularnewline
\hline 
Multiplex Molecular Graph Neural Network (MXMNet) with batch size(BS)
=128 and cut off distance dg \cite{32} & Quantum Chemical & mu, alpha,HOMO ,LUMO, Gap, R2, ZPVE, U0, U, H, G, Cv,Omega, & MAE(avg) & 0.92\% & GNN\tabularnewline
\hline 
\multirow{4}{*}{MT-CGCNN \cite{46} } & \multirow{4}{*}{Electronic} & Multi-task Learning (Formation Energy and Bandgap)  & \multirow{4}{*}{\makecell{Improvement \\ over \\ CGCNN
}} & 8.30\% & \multirow{4}{*}{GNN}\tabularnewline
\cline{3-3} \cline{5-5} 
 &  & (Formation Energy and Fermi Energy) &  & 3.80\% & \tabularnewline
\cline{3-3} \cline{5-5} 
 &  & (Bandgap and Fermi Energy) &  & 1.70\% & \tabularnewline
\cline{3-3} \cline{5-5} 
 &  & (Formation Energy, Bandgap, Fermi Energy) 
 &  & 4.40\%
 & \tabularnewline
\hline  
\multirow{5}{*}{\makecell{Orbital Graph \\ Convolutional \\ Neural Network \\ (OGCNN) \cite{39}}} & \multirow{5}{*}{Electrical} & Lanthanides-formation energy   & \multirow{5}{*}{MAE} & 0.06 ev/atom & \multirow{5}{*}{GNN}\tabularnewline
\cline{3-3} \cline{5-5} 
 &  & Perovskites-formation energy  &  & 0.05 ev/atom & \tabularnewline
\cline{3-3} \cline{5-5} 
 &  & MP-formation energy &  & 0.03 ev/atom & \tabularnewline
\cline{3-3} \cline{5-5} 
 &  & MP-Band Gap &  & 0.032 ev & \tabularnewline
\cline{3-3} \cline{5-5} 
 &  & MP-Fermi Energy &  & 0.38 ev & \tabularnewline
\hline 
OrbNet \cite{112} & Quantum Mechanical & Total Energy for Molecules and Relative Conformer Energy for Molecules &  & 33\% (improvemet over best prior model) Similar accuracy as DFT method & GNN\tabularnewline
\hline 
\multirow{2}{*}{\makecell{ Roost  \\ {[}Ensemble{]} \cite{105}}} & \multirow{2}{*}{Electronic} & \multirow{2}{*}{bandgap} & MAE  & 0.0241 ev & \multirow{2}{*}{GAN}\tabularnewline
\cline{4-5} \cline{5-5} 
 &  &  & RMSE  & 0.0871 ev & \tabularnewline
\hline 
 SOAP + SchNet Model \cite{72} & Electronic & Bandgap & MAE & 0.388 ev & Kernel Ridge Regression + CNN\tabularnewline
\hline 
UNet+proto-DenseNet \cite{91} & To find refined pattern & \_\_\_\_\_\_\_ &  & 80.05\% & CNN\tabularnewline
\hline 
\end{longtable}
\begin{longtable}{|p{2.5cm}|p{2.5cm}|p{6cm}|p{6cm}| } 
%\rowcolors{1}{gray!40!white}{blue!40!white!80} 
\pagebreak
\caption{Dataset details: This table contains the publicly available datasets (both experimental based and DFT based) available in the field of materials science and engineering. The description column gives details about the databases and the last column provides the respective link of that dataset.}
\label{table3}\\[0.5ex]
\hline

   \textbf{Category} &	\textbf{Database Name} &	\textbf{Description} &	\textbf{Link}
 \\[1ex]
 \hline
 Quantum Physical &
 JARVIS-DFT 3D \cite{18} &	The database contains electrical, magnetic and electro-magnetic properties of 40000 bulk and 1000 low-dimensional crystalline materials	& "https://www.nist.gov/programs-projects/jarvis-dft" \\[0.5ex]
\hline
Structural &	Open Quantum Materials Database (OQMD)\cite{18} &	Contains thermodynamic and structural properties of 1,022,603 materials	& https://oqmd.org/ \\[0.5ex]
\hline
Mechanical and Thermal	& (AFLOWLIB) \cite{13}	& The database consists of 3,528,653 material compounds with over 733,959,824 electric, elastic and thermal properties	& http://www.aflowlib.org \\[0.5ex]
\hline
Chemical & Zinc \cite{72} & A collection of commercially available chemical compounds prepared for virtual screening & https://zinc12.docking.org/\\[0.5ex]
\hline
Crystallo-graphic &	Inorganic Crystal Structure Database (ICSD) \cite{73} &	Contains 262242 crystal structures of elements to Quaternary compounds as of May 2022	& "https://icsd.products.
fiz-karlsruhe.de/" \\[0.1ex]
\hline
Structural &	NOMAD \cite{74}	& Contains ab initio electronic-structure data from DFT and other methods	& "https://nomad-lab.eu/index.
php?page=repo-arch" \\[2ex]
\hline
Computational Chemistry	& ioChem-BD \cite{75} & A digital repository that deals with computational chemisty dataset & {https://www.iochem-bd.org/} \\[0.5ex]

\hline
Multivariate	&  Materials Experiment and Analysis Database (MEAD)\cite{73} & "Contains raw and metadata obtained from material synthesis and characterization
experiments also contains the property and performance analysis of that data." & https://solarfuelshub.org/materials-experiment-and-analysis-database  \\[0.5ex]
\hline
Multivariate & UCI Machine Learning Repository \cite{83}	& Is a collection of over 550 datasets	& https://paperswithcode.com/dataset/uci-machine-learning-repository  \\[0.5ex]
\hline
Structural and surface morphology & Crystalium \cite{40} & Contains surface properties of 145 crystals of 74 elements available and Grain Boundary properties of 58 crystals of 58 elements	& http://crystalium.materialsvirtuallab.org/ \\[0.3ex]
\hline
Quantum Physical	& Materials Project(MP)\cite{12} &	It is consist of inorganic compounds, band structures,molecules, nano porous materials and their properties like Magnetic moment, formation energy, energy above hull etc.	& https://materialsproject.org \\[0.5ex]
\hline
Structural	& Crystallography Open Database (COD)\cite{10} & 	an Open-access database containing crystal structures of organic, inorganic, metal-organic compounds and minerals	&  http://www.crystallography.net/cod/ \\[0.1ex]
\hline
Thermo-dyanamics & 	SGTE Solid SUBstance (SSUB) \cite{9}	& contains thermochemical property for about 4300 condensed or gaseous species	& https://www.sgte.net/en/neu \\[0.5ex]
\hline
Quantum Chemistry	& QMOF\cite{87}	& Contains quantum-chemical properties of MOF	& https://github.com/arosen93/QMOF \\[0.5ex]
\hline
Metal–organic framework	& Reduced\_HMOF \cite{87}	& Contains 51,163 unique hypothetical MOFs with genetic information	& https://mof.tech.northwestern.edu/ \\[0.5ex]
\hline
Metal–organic framework	& CoRE MOF-2019 \cite{84}	& Contains 3D porous MOFs that are directly usable in molecular simulations or electronic structure calculations &	https://zenodo.org/record/3370144-\#.Yr3UBnZBzIU \\[0.5ex]
\hline
Energy Material & 	Hydrogen Storage Materials Database \cite{17}	& It contains 2722 hydrogen storage materials with their composition and hydrogen gravimetric capacity	& http://surl.li/cejcd \\[0.5ex]
\hline
Structural	& Cambridge Structure Database [\cite{15}]	&  Contains one million 3-D structural data of molecules	& https://www.ccdc.cam.ac.uk/solutions/csd-core/components/csd/ \\[0.5ex]
\hline
Structural and Quantum Mechanical &	NIMS (MatNavi) Materials database \cite{129}	& Contains properties of Polymer, Inorganic and Matellic materials and computational electronic structures of materials	& https://mits.nims.go.jp/en/ \\[0.5ex]
\hline
Quantum Chemistry & 	QM7\cite{36} & 	A subset of GDB-13 dataset containing 7165 molecules & 	http://quantum-machine.org/datasets/ \\[0.5ex]
\hline
Quantum Chemistry	& QM7b\cite{36}	& An extension of QM7 dataset by predicting 13 additional properties.	& http://quantum-machine.org/datasets/ \\[0.5ex]
\hline
Quantum Chemistry	& QM8 \cite{36}	& Contains electronic spectra and excited space energy of small molecules	& https://moleculenet.org/datasets-1 \\[0.5ex]
\hline
Quantum Chemistry	 & QM9 \cite{71}	& Gives Quantum Chemical Properties of small organic molecules &	http://quantum-machine.org/datasets/ \\[0.5ex]
\hline
Quantum Chemistry	& Free Solvation Database (FreeSolv) 
 \cite{36}	& Contains hydration free energy of experimental and calculated water solubility data	& https://github.com/MobleyLab/FreeSolv \\[0.5ex]
\hline
Quantum Chemistry	& ESOL\cite{36}	& Solubility database of 1128 compounds	& https://integbio.jp/dbcatalog/en/record-/nbdc00440  \\[0.5ex]
\hline
Quantum Chemistry 	& Organic Materials Database (OMDB)\cite{72}	&  Contains electronic properties of organic crystal structures	& https://omdb.mathub.io/dataset. \\[0.5ex]
\hline
XRD	 & DiffraNet \cite{90} &	Contains 25,000 labeled serial crystallography diffraction images.	& https://arturluis.github.io/diffranet/ \\[0.5ex]
\hline
Molecular	& PubChem \cite{36}	& Contains chemical molecules and their activities against biological assays	& https://pubchem.ncbi.nlm.nih.gov/ \\[0.5ex]
\hline
Quantum physical 	& OC20 \cite{64}	& Contains 1,281,040 DFT relaxations for materials, surfaces and adsorbates and their molecular dynamics, randomly perturbed and electronic structure analyses. &	https://github.com/Open-Catalyst-Project/ocp/blob/main/DATASET.md \\[0.5ex]
\hline
Structural and Surface morphology &	Warwick Electron Microscopy Datasets \cite{93}	& Contains 19769 experimental scanning transmission electron microscopy2 (STEM) images, 17266 experimental transmission electron microscopy2 (TEM) images and 98340 simulated TEM exit wavefunctions in three different datasets.	& https://github.com/Jeffrey-Ede/datasets \\[0.5ex]
\hline
Solid-state physics and Synthesis	& Text-mined dataset of inorganic materials synthesis recipes \cite{45}	& Contains 19,488 synthesis entries from 53,538 solid-state synthesis paragraphs that uses using text mining  and natural language processing approaches	& https://figshare.com/articles/dataset/solid-state\_dataset\_2019-06-27\_upd\_json/9722159/3 \\[0.5ex]
\hline
Molecular	& GDB databases \cite{36} 	& Contains small organic molecules up to 13 atoms of C, N, O, S and Cl based on simple chemical stability and synthetic feasibility rules	& https://pubs.acs.org/doi/10.1021/ci600423u \\[0.5ex]
\hline
Structural and physical	& Concrete Compressive Strength \cite{36}	& A multivariate dataset containing concrete compressive strength (MPa) for a given mixture for a particular time was determined from laboratory.	& http://surl.li/cejcc \\[0.5ex]
\hline
Physiology and molecular medicine	& Clintox \cite{36}	& For a total of 1491 drug compounds, clinical trial toxicity (or absence of toxicity) FDA approval status are included.	& https://lifesci.dgl.ai/api/data.html \\[0.5ex]
\hline
Molecular	& LIT-PCBA \cite{89} &	From PubChem dataset, 149 dose-response bioassays are included by removing false positives and assay artifacts but keepings ative and inactive compounds having similar molecular property.	& https://drugdesign.unistra.fr/LIT-PCBA/ \\[0.5ex]
\hline
Bio Molecular	& PDBBind database \cite{36}	& Contains 23496 biomolecular complexes with their binding affinity data	& http://www.pdbbind.org.cn/ \\[0.5ex]
\hline
Reactions 	& Reaxys database\cite{69}	& It contains organic and organometallic reactions	& https://www.reaxys.com/\#/ \\[0.5ex]
\hline
Structural and Surface morphology	& JARVIS\_STM \cite{8}	& Contains scanning tunneling microscope (STM) images	& https://jarvis.nist.gov/login?next=/jarvisstm/
\\[0.5ex]
\hline
Structural	& Database of Wannier tight binding Hamiltonians (WTBH) \cite{86}	& Electronic band structure calculations of WTBH	& https://github.com/usnistgov/jarvis \\[0.5ex]
\hline
Chemical 	& InfoChem\cite{40}	& Contains large number of known reactions and molecules &	https://www.deepmatter.io/about-us/infochem \\[0.5ex]
\hline
Chemical	& Citrination\cite{Laugier}	& An experimental based dataset containing chemical information of materials	& https://citrination.com/datasets \\[0.5ex]
\hline
Physical Chemistry	& Lipophilicity\cite{89}	& Contains chemical structure (SMILES) of 1,130 organic compounds and their n-octanol/buffer solution distribution coefficients at pH 7.4	 & https://deepai.org/dataset/lipophilicity \\[0.5ex]
\hline
BioPhysics	& Maximum Unbiased Validation (MUV)\cite{36} & 	A dataset selected from PubChem BioAssay that contains 17 different tasks of 90 thousand compounds	& https://www.tubraunschweig.de/en/pharm-chem/forschung/baumann/translate-to-english-muv \\[0.5ex]
\hline
BioPhysics	& HIV\cite{36}	& Contains 40000 compounds that have the ability to inhibit HIV replication	& https://data.unicef.org/resources/dataset/hiv-aids-statistical-tables/ \\[0.5ex]
\hline
Physiology	& The Blood–brain barrier penetration (BBBP)\cite{36}	& A blood-brain barrier dataset for prediction of barrier permeability	 & https://github.com/theochem/B3DB \\[0.5ex]
\hline
BioPhysics 	& BACE \cite{36}	& Provides binding results for human b-secretase 1 inhibitors	& https://enamine.net/compound-libraries/targeted-libraries/bace-library
\\[0.5ex]
\hline
Physiology 	& Tox21 \cite{36} & Measures toxicity of 8014 compounds	& https://paperswithcode.com/dataset/tox21-1 \\[0.5ex]
\hline
Physiology &	ToxCast \cite{36}	& Contains toxicity of compounds for and larger in size as compared to Tox21	& https://tox21.gov/data-and-tools/ \\[0.5ex]
\hline
Physiology	& Cider\cite{36}	& Contains market drugs and their adverse reactions	& http://sideeffects.embl.de/\\[0.5ex]
\hline
Molecular Dynamics 	& COLL Dataset \cite{95}	& Contains Molecular Collision configurations	& https://figshare.com/articles/dataset/COLL-\_Dataset\_v1\_2/13289165/1 \\[0.5ex]
\hline
 Crystallography and Spectroscopy &	RRUFF \cite{97}	& Contains Raman spectra, X-ray diffraction	\& chemistry data for minerals & https://rruff.info/
\\[0.5ex]
\hline
Chemical & Melting Point Dataset\cite{101}	& contains melting point of around 8000 chemical structures. 	& https://old.datahub.io/dataset/open-melting-point-data\\[0.5ex]
\hline
Multivariate & Superconductivity Dataset \cite{101}	& Contains Chemical Formula and relevant features of about 21263 superconductors. & https://archive.ics.uci.edu/ml/datasets/super-conductivity+data \\[0.5ex]
\hline
Thermoelectric 	& UCSB Thermoelectrics dataset \cite{101}	& The dataset contains Electrical conductivity, Power factor and Seebeck coefficient of thermoelectric materials	 & https://citrination.com/datasets/150557/sh-ow\_files \\[0.5ex]
\hline
Electronic	& dataset of Strehlow and Cook \cite{101} &	Contains energy bandgaps of semiconductors and Insulators	& https://srd.nist.gov/jpcrdreprint/1.3253115-.pdf \\[0.5ex]
\hline

\end{longtable}

\begin{longtable}{|M{2.8cm}|M{1.5cm}|M{6cm}|M{7cm}| } 
%\rowcolors{1}{gray!40!white}{blue!40!white!80} 
\caption{The table contains the work done in the field of five different types of materials, Inorganic, Organic, Energy-storage,drug and Pharmaceutical and Bio materials. The frameworks used in the fields along with a brief description of the work are also mentioned in the table label}
\label{table4}\\[0.5ex]
\hline 
\multicolumn{4}{| c |}{Begin of Table}\\
 \hline
Material Type & Reference & Model/Framework Name & Description
 \\[0.5ex]
\hline 
 \endfirsthead
\hline
 \multicolumn{4}{|c|}{Continuation of Table \ref{table4}} \\
 \hline 
%Model Type & Property Type & Properties to Predict & \multicolumn{2}{c|}{Accuracy/Performance} & Models based on Techniques\tabularnewline
\hline 
\hline
\endhead

\hline
\endfoot
\hline
\multicolumn{4}{| c |}{End of Table}\\
\hline\hline
\endlastfoot
\multirow{3}{*}{\makecell{Inorganic \\ Materials}} & \cite{55} & Fourier Transformed Crystal Properties (FTCP) & The model is used to predict the structure and chemistry of inorganic
crystals for some targeted property\tabularnewline
\cline{2-4} \cline{3-4} \cline{4-4} 
 & \cite{54} & Modified CGCNN & Local energy prediction of inorganic materials is performed using
Graph Neural Network\tabularnewline
\cline{2-4} \cline{3-4} \cline{4-4} 
 & \cite{92} & Convolutional Neural Network + Deep Dense Network & Artificial Intelligence based prediction of Space Group Determination
problem for powdered XRD pattern of inorganic non-magnetic materials
is carried out.\tabularnewline
\cline{2-4} \cline{3-4} \cline{4-4} 
 & \cite{54} & CGCNN &  CGCNN model is used in inorganic crystals of different compositions
and different structures for finding elemental and local environment
similarities\tabularnewline
\cline{2-4} \cline{3-4} \cline{4-4} 
 & \cite{63} & Conditional GAN and Conditional VAE & Inverse design framework is used to generate inorganic compositions
from desired property, without crystal structure informaton\tabularnewline
\cline{2-4} \cline{3-4} \cline{4-4} 
 & \cite{22} & Random Forest & Random Forest ML algorithms are used for finding critical temperature
of inorganic crystals\tabularnewline
\cline{2-4} \cline{3-4} \cline{4-4} 
 & \cite{45} & Artificial Neural Network and Gaussian Processes & For finding a relationship between reaction condition and resulting
spectra of inorganic nano particles, Artificial Neural Network and
Gaussian Processes have been used.\tabularnewline
\cline{2-4} \cline{3-4} \cline{4-4} 
 & \cite{108} & Generative Machine Learning Model(MatGAN) & A GAN is developed for learning the chemical compositions of inorganic
materials\tabularnewline
\cline{2-4} \cline{3-4} \cline{4-4} 
 & \cite{103} & IRNET & Residual connection is used for predicting formation enthalpy of inorganic
materials\tabularnewline
\hline 
\hline 
\multirow{8}{*}{Organic Materials} & \cite{111} & ORGANIC & A framework named as ORGANIC , which is a Objective-reinforced Generative
model is designed and applied in organic photovoltaic materials.\tabularnewline
\cline{2-4} \cline{3-4} \cline{4-4} 
 & \cite{15} & ALIGNN & The ALIGNN Model is applied in Metal-Organic Freamework (MOF) materials
to screen the MOFs that helps in reduction of CO2\tabularnewline
\cline{2-4} \cline{3-4} \cline{4-4} 
 & \cite{87} & Gaussian Process Regression, Support Vector Regression, Neural Network & Different ML approaches are explored to find out the relationship
of structure of MOF and their methane uptake\tabularnewline
\cline{2-4} \cline{3-4} \cline{4-4} 
 & \cite{66} & Preffered Potential (PFP) & A neural network potential is developed and applied in MOF for molecular
adsobption\tabularnewline
\cline{2-4} \cline{3-4} \cline{4-4} 
 & \cite{112} & Gaussian Process Regression (GPR) & An uncertainty prediction scheme for ML models is developed and tested
in chemical shielding of H NMR in organic crystals and formation energy
prediction of small organic molecules.\tabularnewline
\cline{2-4} \cline{3-4} \cline{4-4} 
 & \cite{113} & SOAP model with kernel ridge regression and Schnet & SOAP kernel and SchNet models are used for bandgap prediction of  crystal
structures of Large Organic molecules\tabularnewline
\cline{2-4} \cline{3-4} \cline{4-4} 
 & \cite{115} & Multi-task Deep Artificial Neural Network & ML model is used for prediction of electronic ground state property
and excited state properties of organic molecules\tabularnewline
\cline{2-4} \cline{3-4} \cline{4-4} 
 & \cite{116} & ChemMIxNet & ChemMIxNet model is developed and applied on water solubility dataset
of small organic molecules\tabularnewline
\hline 
\hline 
\multirow{6}{*}{\makecell{Energy Storage \\ Materials}} & \cite{116} & ChemMIxNet & A ML Model CheMixNet  is used to predict HOMO  value of Organic photovoltaic
cells\tabularnewline
\cline{2-4} \cline{3-4} \cline{4-4} 
 & \cite{49} & MegNet & Designed a GNN named as MegNet to compute the atomic force and stress
tensor of unit cell of battery materials.\tabularnewline
\cline{2-4} \cline{3-4} \cline{4-4} 
 & \cite{10} & Gradient Boosting Decision Tree (GBDT) & ML models are used for classifying solar absorber materials based
on spectroscopic limited maximum efficiency (SLME)\tabularnewline
\cline{2-4} \cline{3-4} \cline{4-4} 
 & \cite{26} & Linear Regression,Reduced Error Pruning (REP) tree,Rotation forest+REP
tree, Random Subspace+REP tree & ML model is designed for Bandgap prediction of solar cell materials\tabularnewline
\cline{2-4} \cline{3-4} \cline{4-4} 
 & \cite{17} & Rep Tree, Random Forest Regression and Neural Networks & ML framework is used for predicting high pressure alloys that can
store hydrogen.\tabularnewline
\cline{2-4} \cline{3-4} \cline{4-4} 
 & \cite{78} & Convolutional Neural Network & CNN model is used to recognize mixed dimensional (2D-0D) Formamidinium
Bismuth Iodides as they are useful in energy consumption\tabularnewline
\cline{2-4} \cline{3-4} \cline{4-4} 
 & \cite{64} & CGCNN, Schnet, Dimenet++ & Different state-of-the art Graph Neural Networks  are applied on a
catalyst dataset.\tabularnewline
\cline{2-4} \cline{3-4} \cline{4-4} 
 & \cite{117} & Artificial Neural Network as classifier and calculator & Neural Networks are used for lithium-ion battery design that can reduce
the computational time by orders of magnitude.\tabularnewline
\hline 
\hline 
\multirow{6}{*}{\makecell{Drug \& \\ Pharmaceutical \\ Materials}} & \cite{116} & ChemMIxNet & CheMixNet model is applied on the compounds that has the ability to
slow down HIV replication in vitro study\tabularnewline
\cline{2-4} \cline{3-4} \cline{4-4} 
 & \cite{36} & Moleculenet & The ML framework Moleculenet is designed and used on properties of
drug molecules.\tabularnewline
\cline{2-4} \cline{3-4} \cline{4-4} 
 & \cite{118} & Densenet & A generative model ,Densenet is used to design de Novo drug.\tabularnewline
\cline{2-4} \cline{3-4} \cline{4-4} 
 & \cite{35} & KV-PLM & The model named KV-PLM is designed to assist in the discovery of drug.\tabularnewline
\cline{2-4} \cline{3-4} \cline{4-4} 
 & \cite{119} & OrbNet & The deep learning model Orbnet is designed and applied in drug dataset
that showed prediction accuracy same as DFT with reduced computational
cost.\tabularnewline
\cline{2-4} \cline{3-4} \cline{4-4} 
 & \cite{120} & Random Forest Classifier & Random Forest classifer is used for segmenting polymer blends  that
are used for pharmaceutical tablets.\tabularnewline
\hline 
\multirow{8}{*}{Bio Materials} & \cite{121} & K-nearest neighbour, support vector machine(SVM), ANN & ML algorithms are used for phase selection of High Entropy Alloys
(HEA) that is used for medical implant\tabularnewline
\cline{2-4} \cline{3-4} \cline{4-4} 
 & \cite{122} & Linear Regression,polynomial Regression model,SVM with three different
kernel functions, regression tree model,k-nearest neighbour model,
and ANN with backpropagation & Employed ML surrogate models combined with experimental methods for
finding High Entropy Alloy (HEA) for targeted property.\tabularnewline
\cline{2-4} \cline{3-4} \cline{4-4} 
 & \cite{123} & Nine different  Classification Models & With the help of a Genetic algorithm, suitable ML model and desciptors
are selected for Phase classification of HEA\tabularnewline
\cline{2-4} \cline{3-4} \cline{4-4} 
 & \cite{124} & Decision Tree & Decision tree classifiers are used for analysing  Cytotoxicity of
nano-particles based on cell viability\tabularnewline
\cline{2-4} \cline{3-4} \cline{4-4} 
 & \cite{125} & Linear Model, Non-linear Model and Convolutional Neural Network (CNN) & ML algorithms are employed on biological composites to explore the
mechanical properties\tabularnewline
\cline{2-4} \cline{3-4} \cline{4-4} 
 & \cite{126} & Random Forest ensembled with Additive Regression (RF\_AR) & Ensemble methods are used in designing bio glass\tabularnewline
\cline{2-4} \cline{3-4} \cline{4-4} 
 & \cite{127} & Support Vector Machine & Support Vector Machine is used for screening the dielectric polymers\tabularnewline
\cline{2-4} \cline{3-4} \cline{4-4} 
 & \cite{128} & ANN models & surrogate models are used for surface properties such as protein adsoption
prediction and cellular response for biodegradable polymers.\tabularnewline
\hline 
\end{longtable}
%\end{appendices}
\section*{Biography }

Dr. Mohendra Roy is an expert in the area of artificial intelligence and experienced in developing biomedical sensors/devices. He is currently serving as an assistant professor at the school of technology of Pandit Deendayal Energy University (PDEU), India. He was a postdoctoral fellow at Delta-NTU Corporate Laboratory for Cyber-Physical Systems of Nanyang Technological University, Singapore. He received his Ph.D. in electronics and information engineering from Korea University, South Korea. He did his master’s in bioelectronics as well as physics from Tezpur Central University, India. He has published several high-quality research papers in IEEE transactions, Biosensors and Bioelectronics Journal, Sensors and Actuators B, etc.

%Dr. Mohendra Roy is an expert in the area of artificial intelligence and experience in developing biomedical sensors/devices. He is currently serviing as an assistant professor at the school of technology of Pandit Deendayal Energy University, India. He was a postdoctoral fellwo at Delta-NTU corporate laboratory for cyber-physical systems of Nanyang Technological University, Singapore. He received his PhD in Electronics and Information Engineering from the Korea University, South Korea. He did his marsters in Physics and Bio-Electronics from the Tezpur Central University, India. He has published serveral high-quality research papers in IEEE transactions, Biosensors and Bioelectronics Journal, Sensors and Actuators B, etc.

%is an expert in the area of artificial intelligence and experienced in developing biomedical sensors/devices. He is currently an assistant professor at the shcool of technology of Pandit Deendayal Energy University (PDEU), India. He was a postdoctoral fellow at Delta-NTU coorporate laboratory for Cyber-Physical System of Nanyang Technological University, Singapore. He received his PhD in Electronics and Information Engineering from Korea University, South Korea. He did his masters in Physics ad Bio-Electronics from Tezpur Central University, India. He has published several high-quality research papers in IEEE transactions, Biosensors & Bioelectronics Journal, Sensors and Actuators B, etc.
\end{document}